%% file: Shim.tex
\documentclass[letterpaper]{article} 
\usepackage{aaai24}  
\usepackage{times}  
\usepackage{helvet}  
\usepackage{courier}  
\usepackage[hyphens]{url}  
\usepackage{graphicx} 
\usepackage{natbib}  
\usepackage{caption} 
\usepackage{algorithm}
\usepackage{algorithmic}

\usepackage{newfloat}
\usepackage{listings}
\DeclareCaptionStyle{ruled}{labelfont=normalfont,labelsep=colon,strut=off} 
\floatstyle{ruled}
\newfloat{listing}{tb}{lst}{}
\floatname{listing}{Listing}
\usepackage{epsfig}
\usepackage{amsmath}
\usepackage{amssymb}
\usepackage{lipsum} 
\usepackage{multirow}
\usepackage{cite}
\usepackage{capt-of}
\usepackage{cuted}
\usepackage{booktabs}

\usepackage[symbol]{footmisc}

\title{Towards Squeezing-Averse Virtual Try-On via Sequential Deformation}
\author{
    Sang-Heon Shim, 
    Jiwoo Chung, 
    Jae-Pil Heo\thanks{Corresponding author}
}
\affiliations{
    Sungkyunkwan University \\
    \{ekzmwww, wldn0202, jaepilheo\}@skku.edu
}

\begin{document}

\maketitle

\begin{abstract}
    In this paper, we first investigate a visual quality degradation problem observed in recent high-resolution virtual try-on approach.
    The tendency is empirically found that the textures of clothes are squeezed at the sleeve, as visualized in the upper row of Fig.~\ref{fig_intro_res}(a).
    A main reason for the issue arises from a gradient conflict between two popular losses, the Total Variation~(TV) and adversarial losses.
    Specifically, the TV loss aims to disconnect boundaries between the sleeve and torso in a warped clothing mask, whereas the adversarial loss aims to combine between them.
    Such contrary objectives feedback the misaligned gradients to a cascaded appearance flow estimation, resulting in undesirable squeezing artifacts.
    To reduce this, we propose a Sequential Deformation~(SD-VITON) that disentangles the appearance flow prediction layers into TV objective-dominant~(TVOB) layers and a task-coexistence~(TACO) layer.
    Specifically, we coarsely fit the clothes onto a human body via the TVOB layers, and then keep on refining via the TACO layer.
    In addition, the bottom row of Fig.~\ref{fig_intro_res}(a) shows a different type of squeezing artifacts around the waist.
    To address it, we further propose that we first warp the clothes into a tucked-out shirts style, and then partially erase the texture from the warped clothes without hurting the smoothness of the appearance flows.
    Experimental results show that our SD-VITON successfully resolves both types of artifacts and outperforms the baseline methods.
    Source code will be available at https://github.com/SHShim0513/SD-VITON.
\end{abstract}

\begin{figure*}[t!]
    \centering
     \includegraphics[width=\linewidth]{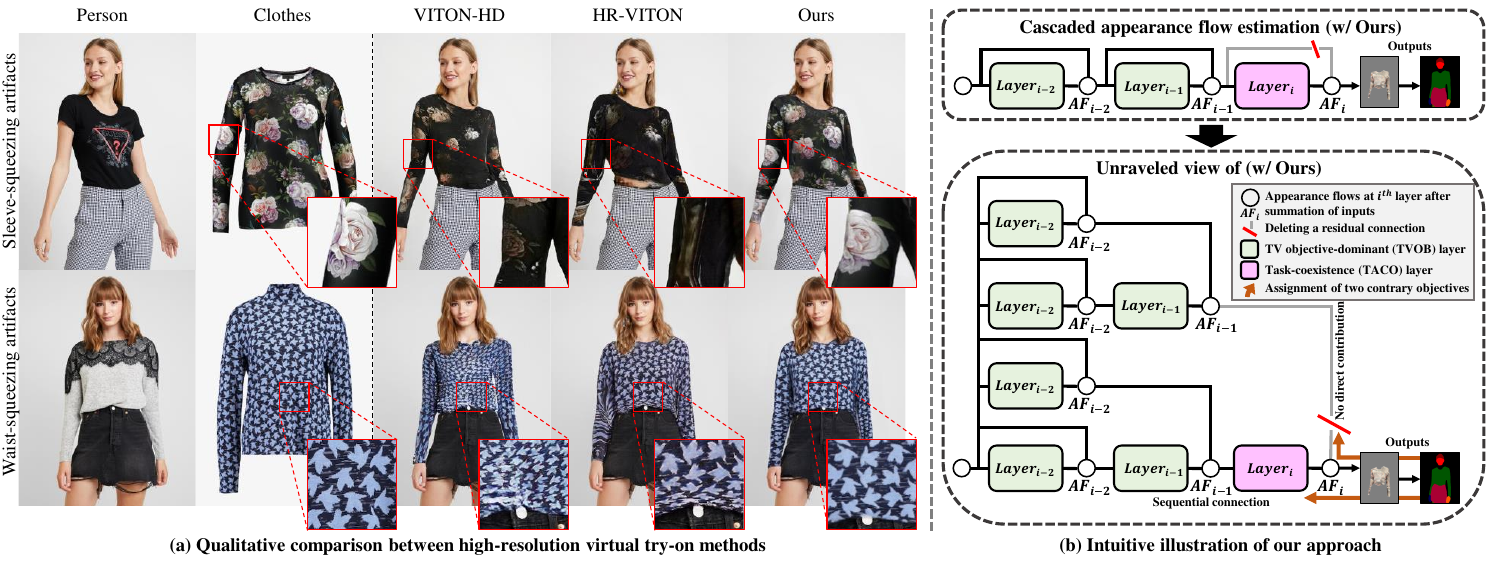}
     \caption{
     (a) Comparing our method with the recent state-of-the-art methods tailored for high-resolution virtual try-on synthesis. On the upper row, HR-VITON~\cite{lee2022hrviton} produces the squeezed clothing textures on the sleeve, while an example at the bottom row shows that HR-VITON generates the squeezed textures around the waist. Those two types of squeezing artifacts are repeatedly observed when the clothes with long sleeves or a person with a tucked-in shirts style is given. Our method achieves significantly better visual quality over the baselines. (b) Intuitive illustration of our approach. Prior methods pass the appearance flows to the following layers via the residual paths, so all layers directly contribute to the warped clothes used in predicting the semantic layout. We disable the passing of previous appearance flows to the last layer by introducing a sequential connection only at the last layer, resulting in the disentanglement of TV objective-dominant layers and a task-coexistence layer.
     }
     \label{fig_intro_res}
\end{figure*}

\section{Introduction}
\label{sec_intro}

The goal of image-based virtual try-on~(VITON)~\cite{han2018viton} is to fit the clothes onto a person.
A general approach is to synthesize a sample through three tasks, which are layout generation, clothes deformation, and full-body image generation.
Its recent advances have reached to synthesize high resolution results with convincing qualities.

A major challenge in high-resolution synthesis is the misalignment between the layout and the warped clothes, since it leads unrealistic textures and is especially more serious in higher resolutions, as shown in~\cite{choi2021viton}.
The latest and state-of-the-art method, HR-VITON~\cite{lee2022hrviton}, greatly alleviates the misalignment problem by introducing a multi-task learning (MTL) approach for virtual try-on. 
Specifically, HR-VITON fuses the layout generation and clothes deformation tasks into a shared backbone with different prediction layers, which can be regarded as MTL.

Nonetheless, repeated failures have been observed in samples with long sleeves, as shown in the upper row of Fig.~\ref{fig_intro_res}(a).
In more detail, a recent VITON approach~\cite{lee2022hrviton} generates virtual try-on samples with squeezed textures at sleeves when the clothes have long sleeves. 
We denote it as a \textit{sleeve-squeezing artifact} throughout this paper.
Since there are many clothes with long sleeves, this issue can disturb the utilization of VITON in real-world applications.

The sleeve-squeezing artifact is mainly caused by the integration of two different tasks into a single model.
In MTL, a task conflict problem~\cite{shi2023recon} is widely known that a model is prone to converge toward a dominant task when the objectives of different tasks are misaligned.
We observed that there is a conflict between two popular losses, the Total Variation~(TV) loss for clothes deformation and the adversarial loss for layout generation.
TV loss aims to smooth the deformation field to preserve the texture of clothes after warping, while adversarial loss aims to generate a layout with a realistic shape according to segmentation maps of training data. Thus, both losses are essential for each task. However, the TV loss leads to disconnect the boundaries between the sleeve and torso in a warped clothes mask, since they are separated in clothes with long sleeves. On the other hand, the adversarial loss for layout generation tries to produce a layout with a connected boundary between the sleeve and the torso following the training data distribution. Since the warped clothes feed into a prediction layer of layout generation, the adversarial loss in layout generation also affects the clothes deformation.

Our raised artifact becomes more serious in a cascaded appearance flow estimation~\cite{ge2021parser,zhou2016view}.
Specifically, the clothes are deformed by appearance flows that are a set of 2D coordinate vectors, where each vector denotes which pixel in clothes image is used to fill a pixel in person image.
In a multi-scale scheme, since the appearance flows are inferred and then accumulated in the subsequent layers through residual paths, all appearance flow prediction layers directly contribute to the warped clothes that are fed to the prediction layer for layout generation.
As a result, all appearance flow prediction layers should struggle to meet the contrary objectives.
In the end, the network converges toward connecting the boundary between the sleeve and the torso, while squeezing the texture of the clothes.

To address the aforementioned problem, we propose a Sequential Deformation~(SD-VITON) that disentangles the appearance flow prediction layers into TV objective-dominant~(TVOB) layers and a task-coexistence~(TACO) layer by disconnecting a direct contribution of intermediate layers to the warped clothes, as illustrated on Fig.~\ref{fig_intro_res}(b). 
Different from the prior method~\cite{lee2022hrviton} where coarse and fine-detailed clothes deformations are done across layers that suffer from contrary goals, our SD-VITON first coarsely fits the clothes onto a human body using the TVOB layers, and then refines minor revisions via the TACO layer which is placed in the end.
By doing this, the TACO layer mainly infers the appearance flows with a small scale, thereby have no enough magnitude of TV loss to cause the artifact.

In addition, the bottom row in Fig.~\ref{fig_intro_res}(a) shows another type of squeezing artifacts around the waist.
We denote it as a \textit{waist-squeezing artifact}.
The waist-squeezing artifacts often occur when a person with a tucked-in shirts style is given.
In this case, it is desirable to eliminate partial contents of the clothes in the warped clothes.
However, prior works choose to preserve whole contents with squeezed textures.
This is mainly because the regularization losses~($e.g.$ TV loss) aim to transform the nearest points in source image into a neighborhood in a target image, thereby a removal of partial contents is not allowed.

To address it, we further propose an approach that we first deform the clothes into a tucked-out shirts style, and then keep on warping into a desired shape by erasing the partial contents of initial result.
Specifically, we enable to produce the tucked-out shirts style by training an appearance flow estimation with a masked loss that excludes the training errors in the torso region outside of the upper clothes.
Plus, we implement an erasing operation in appearance flow estimation by feeding a 0-filled tensor at $z$-axis and encouraging to borrow a 0-value at the same $(x,y)$ coordinate but a different $z$, when fitting into the desired shape.

Extensive experiments validate that SD-VITON successfully addresses two types of squeezing artifacts, and outperform the state-of-the-art works on a high-resolution dataset.

\section{Related Work}
\label{sec_related_work}

\paragraph{Image-based virtual try-on.}
Many works have developed network architectures for generating realistic virtual try-on samples.
CP-VTON~\cite{wang2018toward} proposed a two-stage approach that first deforms the clothes using Thin-Plate-Spline~(TPS) transformation, followed by rendering a try-on sample.
ACGPN~\cite{yang2020towards} additionally introduced a semantic layout generation module that predicts a layout with respect to a person wearing the clothes.
However, there has been a visual quality degradation problem when a misalignment occurs between the semantic layout and the warped clothes.
This is especially more serious in higher resolutions.
To reduce this, VITON-HD~\cite{choi2021viton} removes the information from misaligned regions and then modulates it with a semantic layout during the full-body synthesis stage.
Recently, HR-VITON~\cite{lee2022hrviton} unified two modules of the clothes deformation and the semantic layout generation to reduce the misalignment and handle the occlusions by body parts.

\paragraph{Regularization for appearance flows.} 
Appearance flows control the warping of clothes per pixel.
Thus, minor mistakes could result in unnatural logos or text.
To warp the clothes without hurting the clothing textures, various constraints for enhancing a co-linearity of neighboring appearance flows have been employed.
For instance, most approaches~\cite{lee2022hrviton, han2019clothflow} utilized a Total Variation~(TV) norm, which minimizes the distances of the appearance flows between neighboring pixels.
On the other hand, \cite{choi2021viton}~and~\cite{ge2021parser} employed a second-order smoothing constraint that minimizes the second derivatives of the appearance flows.
Thus, it aims to make the distance between two neighboring intervals equal.
In most cases, minimizing those constraints and the adversarial loss for layout generation bring orthogonal benefits to a network.
However, we found that the two objectives are not well aligned for clothes with long sleeves.

\paragraph{Residual connection.} Prior work~\cite{veit2016residual} has shown that layers with residual connections do not have a strong dependency on each other, behaving like ensembles of different networks.
In VITON, clothes deformation is done across layers, and there is a residual connection between them.
We found that the residual paths between intermediate layers and a last layer seriously cause the sleeve-squeezing artifact since each layer acts like an independent network so that all layers should struggle to meet contrary objectives.
To reduce this, we remove the residual path between the intermediate layers and the last layer.

\section{Our Approach}
\label{sec_method}

\begin{figure*}[t!]
    \centering
     \includegraphics[width=\linewidth]{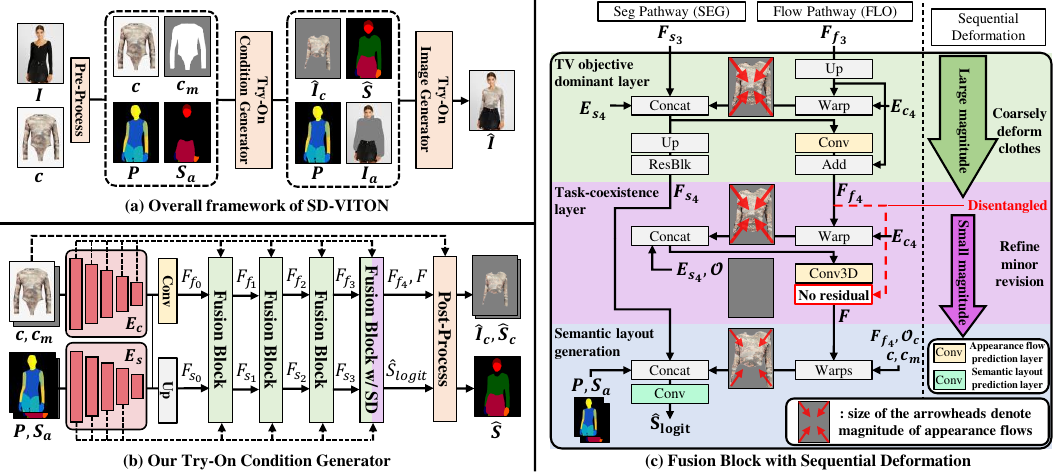}
     \caption{(a) Overall framework of SD-VITON. Our try-on condition generator produces a semantic layout and warped clothes without squeezing artifacts. Our try-on image generator then produces a try-on sample and a composition mask. By employing a composition mask, our SD-VITON generates a final try-on sample. (b) Our try-on condition network. To avoid our raised sleeve- and waist-squeezing artifacts, we propose a fusion block with sequential deformation at the last fusion block. (c) Our fusion block with sequential deformation. To address sleeve-squeezing artifacts, we build an additional appearance flow prediction layer in the flow pathway, and disable the residual connection between them. This results in disentangling the last flow prediction layer and earlier flow prediction layers into a task-coexistence layer and TV objective-dominant layers.
     To address the waist-squeezing artifacts, we then revise the task-coexistence layer to employ an appearance flow estimation with a 0-filled tensor.
     }
     \label{main_figure}
\end{figure*}

Given a person image~($I \in \mathbb{R}^{3\times H\times W}$) and clothes~($c \in \mathbb{R}^{3\times H\times W}$), our goal is to synthesize a try-on sample~($\hat{I} \in \mathbb{R}^{3\times H\times W}$) of $I$ wearing $c$ while preserving the unique texture of $c$.
Since our raised problem occurs in a multi-task learning~(MTL) architecture, we build our method upon the latest MTL based VITON framework~\cite{lee2022hrviton}.

Let us first briefly introduce a procedure for generating $\hat{I}$.
It is composed of three steps as illustrated in Fig.~\ref{main_figure}(a) : (1) a pre-processing; (2) a try-on condition generation; (3) a try-on image generation.
Following the prior works~\cite{choi2021viton, lee2022hrviton}, we first pre-process inputs to obtain a clothing-agnostic person representation~($S_a$ and $I_a$) by removing any clothing information.
Afterwards, we generate a semantic layout~($\hat{S}$) and warped clothes~($\hat{I}_c$) using a try-on condition generator, where $\hat{S}$ denotes a semantic segmentation map with respect to $I$ wearing $c$.
Finally, a try-on image generator is employed to synthesize $\hat{I}$ using $\hat{S}$ and $\hat{I}_c$.

Our pre-processing step is the same as in \cite{lee2022hrviton,choi2021viton}.
We infer a segmentation map ($S \in \mathbb{L}^{H\times W}$) where $\mathbb{L}$ denotes a set of integers that indicate the semantic labels with respect to a human parsing~\cite{gong2018instance}, a densepose map~\cite{guler2018densepose} ($P \in \mathbb{R}^{3 \times H\times W}$), a clothing-agnostic segmentation map~($S_a$), and a clothing-agnostic image~($I_a$) from $I$.
We feed $S_a$, $I_a$, $P$, $c$, and clothes mask~($c_m$) into our networks.

\subsection{Preliminary}
We first explain the MTL architecture of a baseline, HR-VITON, with several notations used throughout the paper.
Please refer to a graphical illustration in Fig.~\ref{main_figure}(b) except for a ``Fusion Block w/ SD" which is main focus of this paper.

\paragraph{Pyramid feature extraction.}
There are a clothing encoder~($E_c$) and a segmentation encoder~($E_s$).
Given ($c$, $c_m$) and ($S_a$, $P$), the baseline first extracts the feature pyramids $\{E_{c_l}\}_{l=0}^{4}$ and $\{E_{s_l}\}_{l=0}^{4}$ from two encoders, respectively, and then feeds them into the fusion blocks.

\paragraph{Fusion block} has two pathways, a segmentation~(SEG) pathway and a flow~(FLO) pathway.
In a $l$-th block, SEG processes the joint feature map~($F_{s_l}$) by taking in $E_{c_l}$ and $E_{s_l}$, while FLO predicts appearance flows~($F_{f_l}$), where $F_{f_l}$ is a set of 2D coordinate vectors.
Each vector denotes which pixel in clothes image are used to fill a pixel in person image.

\paragraph{Cascaded appearance flow estimation.}
To warp $c$ with coarse-to-fine, the $F_{f_{l-1}}$ predicted in earlier layer is summed to $F_{f_{l}}$ in the following fusion block as following recurrence:
\begin{eqnarray}
    & y = F_{f_{l}} + \mathcal{U}(F_{f_{l-1}}),
    \label{eq_recursive}
\end{eqnarray}
where $\mathcal{U}$ is a spatially up-sizing operator. 

\paragraph{Outputs.} At the last, $4$-th, fusion block, the semantic layout~($\hat{S}$) and warped clothes~($W(c,F_{f_{4}})$) are inferred. Specifically, $c$ is warped via a grid-sampling operator~($\mathcal{S}$) with $F_{f_{4}}$:
\begin{eqnarray}
     & W(c,F_{f_{4}}) = \mathcal{S}\big(c, F_{f_{4}} \big).
\end{eqnarray}
In addition, the logits of semantic layout~($\hat{S}_{\text{logit}}$) are inferred;
\begin{eqnarray}
    \hat{S}_{\text{logit}} = \text{conv}_{\text{2d}}\big( [ P, S_a, W(c,F_{f_{4}}), W(c_m,F_{f_{4}}), F_{s_4} ] \big),
    \label{eq_layout}
\end{eqnarray}
where $\text{conv}_{\text{2d}}(\cdot)$ is a convolution operation with a $3\times 3$ kernel, and $[\cdot]$ is a tensor concatenation operator across the channel axis. After a post-process~\cite{lee2022hrviton}, a warped clothes~($\hat{I}_c$) and a semantic layout~($\hat{S}$) is finally produced.

\subsection{Problem Definition}
Total Variation~(TV) loss aims to preserve the texture of $c$ during warping.
Given $F_{f_l}$ inferred from $l$-th layer, it is done by smoothing the neighboring coordinate vectors as follows:
\begin{eqnarray}
    & \mathcal{L}_{\text{TV}} = \sum\limits_{p} \sum\limits_{\pi \in N_p^{tv}} ||F_{f_l}^{p+\pi} - F_{f_l}^p ||_1 ,
    \label{eq_tv}
\end{eqnarray}
where $F_{f_l}^p$ indicates a 2D coordinate vector at the $p$-th position in $F_{f_l}$, and $N_p^{tv}$ is a set of index displacements used to denote the vertical and horizontal neighbors of $p$.
On the other hand, the adversarial loss aims to generate a realistic $\hat{S}$, given $c$, $c_m$, $S_a$, and $P$, and is trained with a discriminator.

Although both losses generally help to synthesize a realistic $\hat{S}$ and $\hat{I}_c$, they can cause a conflicting gradient problem~\cite{shi2023recon} when $c$ with long sleeves is given. This is because the boundaries between the arm and torso are disconnected in $c_m$ with long sleeves, whereas there are many real layouts where the boundaries between them are connected.
Specifically, Eq.~\ref{eq_tv} leads $\hat{I}_c$ toward following the boundaries alike $c_m$ and the adversarial loss guides $\hat{S}$ to follow the real distribution of layout, so both losses are not well aligned with long sleeves. 
Concretely, the conflict arises from the FLO of fusion blocks.
Since $W(c,F_{f_4})$ and $W(c_m,F_{f_4})$ fed into Eq.~\ref{eq_layout} provide cues for predicting the upper clothing regions of $\hat{S}_{\text{logit}}$, a significant adversarial loss is induced when $W(c,F_{f_4})$ and $W(c_m,F_{f_4})$ have a disconnected boundary between the arm and torso due to TV loss.

The raised problem becomes more serious in a cascaded appearance flow estimation.
Since $F_{f_4}$ is obtained by a recursive equation according to Eq.~\ref{eq_recursive}, all appearance flow prediction layers of the FLO directly contribute to $W(c,F_{f_4})$.
Concretely, since $F_{f_4}$ was accumulated by proceeding flow prediction layers of the FLO through residual paths, the objective of layout generation is forcefully assigned to all layers of the FLO.
As a result, the TV and adversarial losses are in conflict across prediction layers of the FLO.
In most cases, it is empirically observed that the TV loss is dominated by the adversarial loss at the boundary region between the arm and torso, resulting in sleeve-squeezing artifacts.

\subsection{Fusion Block with Sequential Deformation}
The aforementioned issue motivates us to propose a Sequential Deformation~(SD) in a try-on condition generator.
As illustrated in Fig.~\ref{main_figure}(b), we introduce our SD into a last fusion block to alleviate the squeezing artifacts.

\paragraph{Sequential connection.} 
We develop to disentangle a cascaded appearance flow estimation into TV objective-dominant~(TVOB) layers and a task-coexistence~(TACO) layer.
To reduce artifacts in the sleeves, we first coarsely fit $c$ into $I$ by using the TVOB layers, and then refine the minor details via the TACO layer which is placed in the end.
Since the TVOB layers deal with large deformation of $c$ in the early layers, and the TACO layer only predicts fine-detailed appearance flows with small scale, there is no enough magnitude of TV loss to cause artifacts at the TACO layer.

To disentangle this, we introduce two successive prediction layers with a sequential connection within the FLO of last fusion block, as illustrated in Fig.~\ref{main_figure}(c).
Specifically, we prevent the direct contribution of the intermediate prediction layers from computing the warped clothes and the warped clothes mask, which are used to predict $\hat{S}_{\text{logit}}$, by disabling the summing operation between two successive prediction layers in the FLO.
By doing this, the appearance flows predicted by the last prediction layer are disconnected from the preceding appearance flows, thereby the objective of layout generation is mainly considered in the last prediction layer.
This results in disentangling the intermediate prediction layers and the last prediction layer into the TVOB layers and the TACO layer.

Let us describe the practical implementation of our sequential connection.
Our explanation starts with the last fusion block, where $c$ is firstly warped by the TVOB layer, followed by deformation via the TACO layer.
Given the previous appearance flows~($F_{f_{3}} \in \mathbb{R}^{2\times h_{3} \times w_{3}}$), clothes-based feature map~($E_{{c}_{4}}$), a segmentation-based feature map~($E_{{s}_{4}}$), and a joint feature map~($F_{{s}_{4}}$) from the SEG, we first warp $E_{{c}_{4}}$ and predict an initial appearance flows~($F_{f_{4}}$) as follows:
\begin{eqnarray}
    & F_{f_{4}} = \text{conv}_{\text{2d}}\big( [\mathcal{S}\big(E_{{c}_{4}}, \mathcal{U}(F_{f_{3}})\big), F_{{s}_{3}}, E_{{s}_{4}}] \big),
\end{eqnarray}
where $\mathcal{S}$ is a grid-sampling operator, $\mathcal{U}$ is a spatially up-sizing operator, and $[\cdot]$ denotes a tensor concatenation operator across channel axis.
Since it is a TVOB layer, we add the previous $F_{f_{3}}$ to $F_{f_{4}}$ by employing Eq.~\ref{eq_recursive},
\begin{eqnarray}
    & y = F_{f_{4}} + \mathcal{U}(F_{f_{3}}). \nonumber
    \label{eq_initial_warping}
\end{eqnarray}

Let us then explain the TACO layer.
Given the $E_{{c}_{4}}$, $E_{{s}_{4}}$, and $F_{f_{4}}$, we coarsely fit $E_{{c}_{4}}$ onto a human body by utilizing the $F_{f_{4}}$ inferred from TVOB layers. Then, we predict the appearance flows~($F_{\text{2d}} \in \mathbb{R}^{2 \times h_4 \times w_4}$) as follows,
\begin{eqnarray}
    & F_{\text{2d}} = \text{conv}_{\text{2d}}([\mathcal{S}(E_{{c}_{4}}, F_{f_{4}}), E_{{s}_{4}}]).
    \label{eq_sequential_connection}
\end{eqnarray}
However, we do not add $F_{f_{4}}$ to $F_{\text{2d}}$ to implement a sequential connection.
Note that Fig.~\ref{main_figure}(c) displays an integrated implementation of a sequential connection and the removal of the region of no interest, so it differs from Eq.~\ref{eq_sequential_connection}. 

\paragraph{Removal of the region of no interest.} 
We identify that waist-squeezing artifacts mainly occur when $I$ with a tucked-in shirts style is given.
In such cases, it is desirable that partial contents of $c$ should be eliminated in $\hat{I}_c$.
However, we empirically observed that existing VITON frameworks choose to retain the entire contents of $c$ while squeezing the textures in a waist.

This motivates us to propose an approach that we first deform $c$ into a tucked-out shirts style, and then keep on warping into a desired shape by erasing the region of no interest~(non-ROI), where the non-ROI denotes the set of pixels outside the clothing region of $I$.
However, there are two difficulties to realize our idea. First, there is no explicit way to deform $c$ with a tucked-out shirts style. Second, we should erase the non-ROI without hurting the smoothness of the appearance flows. Otherwise, we have identified that waist-squeezing artifacts still remain. 

\begin{figure}[t]
    \centering
     \includegraphics[width=\linewidth]{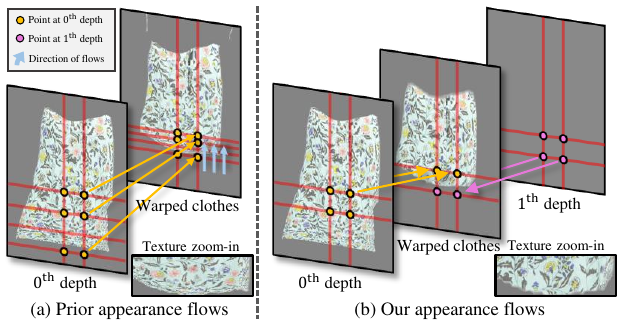}
     \caption{Removal of the non-ROI. Given the warped clothes with tucked-out shirts style, prior appearance flow estimation borrows a 0 value from the lower $y$ coordinate to erase the non-ROI. However, our appearance flow estimation can borrow a 0 value from the same $(x,y)$ coordinate but a different $z$. This results in erasing the non-ROI without hurting the smoothness of appearance flows.} 
     \label{fig_4_temp}
\end{figure}

To deal with these difficulties, we propose a masked loss and an appearance flow estimation with a 0-filled tensor.
Let us first discuss about the masked loss. 
With the densepose map $P$, we first define a mask $\mathcal{M}$ to describe torso regions outside of the upper clothes as follows,
\begin{eqnarray}
    & \mathcal{M} = P_{\text{torso}} \odot (1-S_c),
\end{eqnarray}
where $P_{\text{torso}}$ is a mask representing the torso region, and $S_c$ is a clothing mask for the upper clothes.
To warp $c$ for a tucked-out shirts style, we exclude the regions defined by $\mathcal{M}$ in evaluating training errors. 
Specifically, we train our TVOB layers to minimize the reconstruction-based losses after multiplying $(1-\mathcal{M})$.
The perceptual~\cite{johnson2016perceptual} and L1 losses are minimized using a ground-truth of clothing textures~($I_c$) and clothing mask~($S_c$) as follows,
\begin{eqnarray}
    & \mathcal{L}_{\text{VGG}}^{\mathcal{M}} = \sum\limits_{l=0}^{4}\phi \big( W(c, F_{f_l}) , I_c, (1-\mathcal{M}) \big), \text{ and} \\ 
    & \mathcal{L}_{\text{L1}}^{\mathcal{M}} = \sum\limits_{l=0}^{4}\| \big( W(c_m, F_{f_l}) - S_c \big) \odot (1-\mathcal{M}) \|_1,
\end{eqnarray}
where $\phi(A,B,C)$ denotes a function of a perceptual loss between $A$ and $B$ only for pixels specified in the mask $C$. 
It enables us to warp $c$ down to the end of torso.

Given $W(c,F_{f_4})$ which represents a warped clothes with a tucked-out shirts style, our goal is to erase the non-ROI to fit $c$ into $I$.
However, the appearance flow estimation of Eq.~\ref{eq_sequential_connection} mainly hurts the smoothness of appearance flows while removing the non-ROI, since 2D appearance flow estimation erases texture by sampling a zero value at other $(x,y)$ coordinate of source image, as illustrated in Fig.~\ref{fig_4_temp}.

To address this, our main idea begins with the followings.
If the appearance flow estimation learns to borrow a zero value from the $(x,y)$ coordinate of itself, we can remove the texture in the non-ROI while maintaining the smoothness of appearance flows in the foreground. 
To do this, we expand a dimension, and concatenate a 0-filled tensor~($\mathcal{O}$) along the $z$ axis, where $\mathcal{O}$ has the dimensions of the input of Eq.~\ref{eq_sequential_connection}.
We inform that a warped clothes resides in $0^{\text{th}}$ depth, while $\mathcal{O}$ is in $1^{\text{th}}$ depth.
Then, we predict the appearance flows~($F \in \mathbb{R}^{3 \times 2 \times h_4 \times w_4}$) which represent a set of $(x,y,z)$ coordinate vectors for each pixel and depth in person image as follows,
\begin{eqnarray}
& F = \text{conv}_{\text{3d}}(\big<[\mathcal{S}(E_{{c}_{4}}, F_{f_{4}}), E_{{s}_{4}}], \mathcal{O}\big>),    
\label{eq_af_0_filled}
\end{eqnarray}
where $\text{conv}_{\text{3d}}(\cdot)$ is a convolution operator with a $3\times 3\times 3$ kernel, and $\big< \cdot, \cdot \big>$ is an operator that expands each tensor into a depth dimension before concatenating tensors along the depth axis.
By doing this, the TACO layer can borrow a $0$-value from the same $(x,y)$ coordinate but a different $z$ coordinate when erasing the texture from $W(c,F_{f_4})$ is required. 

Lastly, we create a tensor filled with zeros, having 3 channels~($\mathcal{O}_c \in \mathbb{R}^{3 \times h_c \times w_c}$), where $h_c$ and $w_c$ denote the height and the width of $c$. We deform $c$~(or $c_m$) using $F_{f_4}$ and $F$ as follows,
\begin{eqnarray}
   & W^{0^{\text{th}}}(c,F) = \mathcal{S}\big(\big<\mathcal{S}(c, F_{f_{4}}), \mathcal{O}_c \big>, F \big),
\end{eqnarray}
where $W^{0^{\text{th}}}(c,F)$ indicates a warped clothes at $0^{\text{th}}$ depth, which is a final output of FLO in our fusion block with SD.

In addition, we introduce a distance loss along $z$ axis to guide $F$.
It enforces the borrowing of a zero value from $\mathcal{O}_c$ if the pixel belongs to the non-ROI as follows,
\begin{eqnarray}
    \mathcal{L}_{\text{z-dist}} = \sum\limits_{p \in P_z^{0^{\text{th}}}}(\underbrace{|| (z^{0^{\text{th}}} - F^p) \odot S_c ||_1}_{\text{Region of Interest}} \nonumber \\ + \underbrace{|| (z^{1^{\text{th}}} - F^p ) \odot (1 - S_c) ||_1}_{\text{Region of no Interest}}),
\end{eqnarray}
where $P_z^{0^{\text{th}}}$ is a subset of indices of $F$, whose $z$ coordinate belongs to the $0^{\text{th}}$ depth.
Also, $F^p$ indicate a coordinate vector of $F$ at index $p$.
The $z^{0^{\text{th}}}$ and $z^{1^{\text{th}}}$ are constant scalars corresponding to the ground-truth $z$ coordinates for the $0^{\text{th}}$ and $1^{\text{th}}$ depths. 
They have values of $-1$ and $1$, respectively. 

As our primary focus in this paper lies on the SD, we simply follow the other implementation details of~\cite{lee2022hrviton} incorporating $\mathcal{L}_{\text{CE}}$, $\mathcal{L}_{\text{L1}}$, $\mathcal{L}_{\text{VGG}}$, $\mathcal{L}_{\text{cGAN}}$, $\mathcal{L}_{\text{TV}}$.
As a result, our try-on condition generator is trained as follows,
\begin{eqnarray}
    & \mathcal{L}_{\text{TOCG}} = \lambda_{\text{CE}}\mathcal{L}_{\text{CE}} + \lambda_{\text{L1}}(\mathcal{L}_{\text{L1}} + \mathcal{L}_{\text{L1}}^{\mathcal{M}})  
    \nonumber \\ & + (\mathcal{L}_{\text{VGG}} + \mathcal{L}_{\text{VGG}}^{\mathcal{M}}) + \mathcal{L}_{\text{cGAN}} + \lambda_{\text{TV}}\mathcal{L}_{\text{TV}} + \mathcal{L}_{\text{z-dist}},
\end{eqnarray}
where $\lambda_{\text{CE}}$, $\lambda_{\text{L1}}$, $\lambda_{\text{TV}}$ is set to $10$, $10$, and $2$ for a balance.

\begin{table}[t]
	\centering
	\begin{tabular}{l|cc|cc}
		\toprule
		Methods  & $\text{FID}_\downarrow$ & $\text{KID}_\downarrow$ & $\text{LPIPS}_\downarrow$ & $\text{SSIM}_\uparrow$ \\
		\midrule
		\midrule
        CP-VTON & 43.28 & 3.762 & 0.158 & 0.786 \\
		ACGPN & 43.29 & 3.730 & 0.112 & 0.850 \\
		PF-AFN & 14.01 & 0.588 & N/A & N/A \\
        \midrule
        VITON-HD & 11.59 & 0.247 & 0.077 & 0.873 \\
        HR-VITON & 10.91 & 0.179 & 0.065 & 0.892 \\
        HR-VITON*  & 10.43 & 0.179 & 0.0643 & 0.8919 \\
        Ours & \textbf{9.75} & \textbf{0.120} & \textbf{0.0616} & \textbf{0.8955} \\
		\bottomrule
	\end{tabular}
    \caption{Quantitative results. HR-VITON* denotes HR-VITON with a composition mask, which is the same setting with ours. The KID scores are amplified by 100 times.
    }
	\label{tb_overall_result}
\end{table}

\subsection{Try-On Image Generation}
\label{sec_try_on_image_gen}
Given a clothing-agnostic image~($I_a$), $P$, $\hat{I}_c$, and $\hat{S}$, we generate a try-on sample~($\hat{I}$) via a try-on image generator.
We follow the approach of \cite{lee2022hrviton}, but one different detail is that our try-on image generator additionally outputs a composition mask~($\hat{m}$) employed to produce $\hat{I}$ as follows:
\begin{eqnarray}
    & \hat{I} = \hat{I}_g \odot (1 - \hat{m}) + \hat{I}_c \odot \hat{m},
\end{eqnarray}
where $\hat{I}_g$ is a generated clothed person by the try-on image generator, and $\odot$ denote an element-wise multiplication.

\begin{figure}[t!]
\centering
\includegraphics[width=\linewidth]{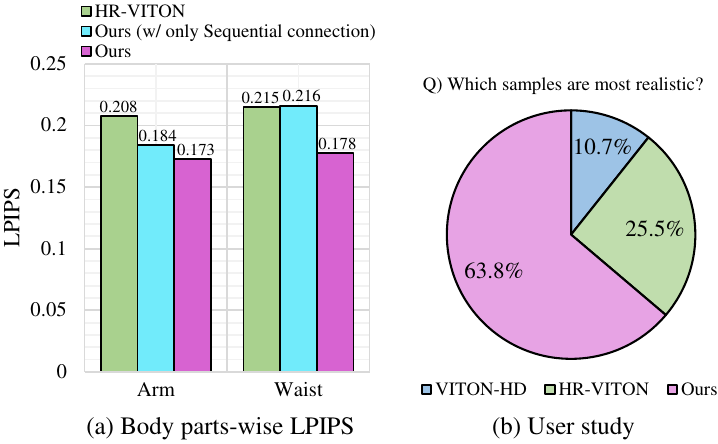}
\caption{Results of body parts-wise LPIPS and user study.}
\label{LPIPS_parts_user_study}
\end{figure}

\section{Experiments}
\label{sec_eval}
\subsection{Experimental Setup}

\paragraph{Dataset.} We conduct experiments on a high-resolution virtual try-on dataset introduced by VITON-HD~\cite{choi2021viton}.
It contains $11,647$ image pairs for the training phase and $2,032$ pairs for the evaluation, each of which has a front-view woman and a top clothes with $1024 \times 768$ resolution.

\paragraph{Baselines.} We evaluate our method with the recent virtual try-on methods including CP-VTON~\cite{wang2018toward}, ACGPN~\cite{yang2020towards}, PF-AFN~\cite{ge2021parser}, VITON-HD~\cite{choi2021viton}, and HR-VITON~\cite{lee2022hrviton}.
We also report HR-VITON with a composition mask, which is the same setting with our try-on image generator.

\paragraph{Evaluation Metrics.} We evaluate the virtual try-on results on both paired and unpaired setting.
In the paired setting, each of methods aims to reconstruct the person image with the original clothing item.
The unpaired setting is to synthesize the person with different top clothing item.
We compute SSIM~\cite{wang2004image} and LPIPS~\cite{zhang2018perceptual} on the paired setting, and compute Frechet Inception Distance~(FID)~\cite{heusel2017gans} and Kernel Inception Distance~(KID)~\cite{bińkowski2018demystifying} on the unpaired setting.

\subsection{Quantitative Results}

\paragraph{Evaluation on try-on sample.} Table.~\ref{tb_overall_result} reports the LPIPS, SSIM, FID, and KID measured from the $2,032$ test pairs.
The results show that our method consistently outperforms all comparing baselines over all evaluation metrics.
Specifically, our method achieves a FID value of $9.75$, improving $0.68$ points, over HR-VITON*, as well as increases the LPIPS value from $0.0643$ to $0.0616$.
Similar trends can be found in SSIM and KID metrics.
Extensive evaluations with diverse metrics clearly validate that our SD-VITON generates the try-on samples much closer to a realistic quality.

\paragraph{Body parts-wise LPIPS on warped clothes.} We then validate that our methods improve the warping quality with respect to a waist and an arm.
To show it, we identified the patches nearby a waist and an arm by employing a densepose and a layout, and then measure the LPIPS of a warped clothes followed by adaptively aggregating the errors per each body parts.
To remove the distracting factors, we manually select $182$ pairs in a test set, each of which are a person with a tucked-in shirts style and the clothes with a long sleeve.
We report the LPIPS of warped clothes with respect to the selected test pairs.
Fig.~\ref{LPIPS_parts_user_study}(a) shows that the LPIPS on the arm is improved when our sequential connection is applied to HR-VITON.
However, the LPIPS nearby the waist did not improve yet.
On the other hand, when the removal of non-ROI is applied upon a model of ours w/ sequential connection, we observed that the errors nearby the waist are further improved.
These results validate that each of our methods improves the warping quality with respect to the specific body parts, and their orthogonal benefits.

\paragraph{User study.}
From the test dataset, we randomly selected 30 sets in unpaired setting. 15 participants were involved, and each user was asked to choose the most realistic sample from the results produced by different methods. Fig.~\ref{LPIPS_parts_user_study}(b) shows that our method achieves the highest selection ratio. 

\begin{figure}[t]
    \centering
    \includegraphics[width=\linewidth]{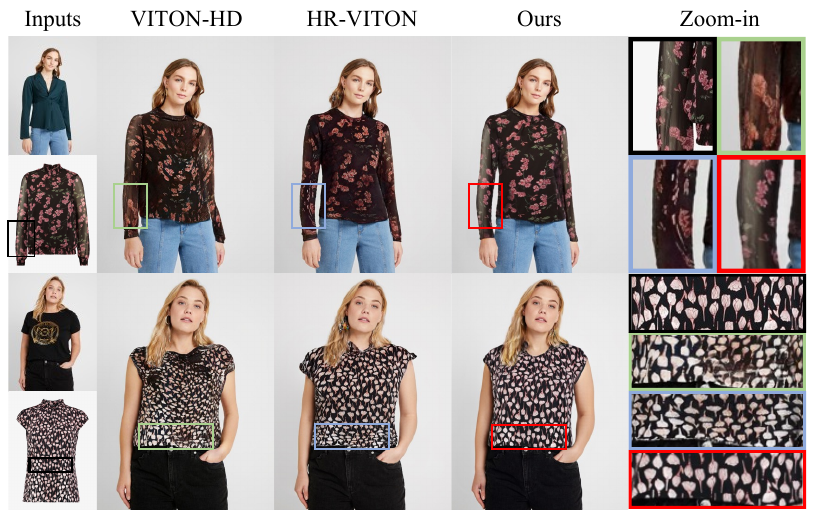}
    \caption{Qualitative comparison. Upper row shows a comparison with respect to the sleeve-squeezing artifact. Bottom row is a comparison related to the waist-squeezing artifact.
    }
    \label{qualitative_twocol_res}
\end{figure}

\subsection{Qualitative Results}

\paragraph{Sleeve-squeezing artifact.} The upper row of Fig.~\ref{qualitative_twocol_res} shows that our method generates the virtual try-on samples in a high quality, when a long sleeve is given.
As seen in the zoom-in, our approach warps the clothes with preserving the flower pattern on the arm, while HR-VITON warps it with a seriously squeezed pattern.
One may found that samples of VITON-HD also do not suffer from the sleeve-squeezing artifact. 
But, overall quality is unrealistic than other methods.

\paragraph{Waist-squeezing artifact.} The bottom row of Fig.~\ref{qualitative_twocol_res} shows that our method also addresses squeezing artifacts around a waist.
The comparison of zoom-in shows that our approach synthesizes the ginkgo leaves with consistent intervals.
On the other hand, the intervals are dramatically narrowed in the sample of HR-VITON.
Those results of Fig.~\ref{qualitative_twocol_res} confirm that our methods can synthesize try-on samples without squeezing artifacts when the clothes with a long sleeve or a person with a tucked-in shirts style is given.

\begin{figure}[t!]
	\centering
    \includegraphics[width=\linewidth]{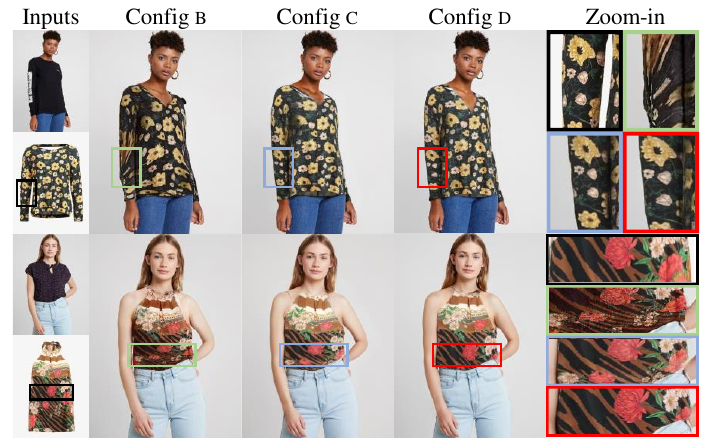}    
    \caption{Qualitative comparison between the ablated models. The upper and bottom rows are related to the sleeve-and waist-squeezing artifacts, respectively.}
	\label{ablation}
\end{figure}

\subsection{Ablation Study}

\paragraph{Employment of a composition mask.} We first revise for the try-on image generator to synthesize the samples with a composition mask.
As reported on Table.~\ref{tb_ablation_study}, HR-VITON* improves the FID scores from $10.91$ to $10.43$.
This is because clothing textures in the warped clothes are not erased by a try-on image generator, which is shown on the second column of Fig.~\ref{ablation}.
However, the employment of a composition mask does not resolve our raised squeezing artifacts.

\begin{table}[t]
\centering
\begin{tabular}{l|cc|cc}
		\toprule
		Configuration & $\text{FID}_\downarrow$ & $\text{KID}_\downarrow$ & $\text{LPIPS}_\downarrow$ & $\text{SSIM}_\uparrow$  \\
		\midrule 
		\midrule
		{\sc a}~ HR-VITON	& 10.91 & 0.179 & 0.065 & 0.892   \\
        {\sc b}~ HR-VITON*  & 10.43 & 0.179 & 0.0643 & 0.8919 \\
        {\sc c}~+~Se.~connection & 10.08 & 0.150 & 0.0623 & 0.8952 \\
        {\sc d}~+~Rem.~non-ROI & \textbf{9.75} & \textbf{0.120} & \textbf{0.0616} & \textbf{0.8955} \\
		\bottomrule
	\end{tabular}
\caption{The ablation study of our method.} 
\label{tb_ablation_study}
\end{table}

\paragraph{Sequential connection.} We then add our sequential connection to HR-VITON*.
As reported in third row, this ablated model achieves the FID of $10.08$, improving $0.35$ points, over HR-VITON*.
The improvements are reasonable since the addition of sequential connection addresses the sleeve-squeezing artifact, thereby much realistic texture of the clothes is synthesized in a sample.
Qualitative results are shown at column of config.~{\sc c}.
However, the ablated model still suffers from the squeezing problem around a waist.

\paragraph{Removal of non-ROI.} Lastly, we integrated a removal of non-ROI upon the config.~{\sc c}.
As reported in the last row, our full model further improves the visual quality, achieving an FID of $9.75$.
Qualitative result shows that the full model further addresses the waist-squeezing artifact.
Those ablation studies confirm that each components of our method are essential for solving squeezing artifacts.

\section{Conclusion}
\label{sec_conclude}

In this paper, we have categorized two squeezing artifacts.
The sleeve-squeezing artifact is caused due to contrary objectives between TV loss and adversarial loss.
On the other hand, the waist-squeezing artifact is caused by prior techniques not able to partially remove the clothing texture.
To reduce them, we coarsely fits the clothes into a person using TV objective-dominant layers, and then refine in the task-coexistence layer.
In addition, we have proposed a removal operation of the partial texture of clothes via appearance flow estimation with a 0-filled tensor.
Experimental results show that our method can resolve both types of artifacts.

\section*{Acknowledgments}
This work was supported in part by MSIT/IITP (No. 2022-0-00680, 2019-0-00421, 2020-0-01821, 2021-0-02068), and MSIT\&KNPA/KIPoT (Police Lab 2.0, No. 210121M06).

\bibliography{aaai24}

\newpage

\twocolumn[
\begin{center}
    \vspace*{1.1cm}
    \LARGE{\bf{Supplementary Material \\ \Large{- Towards Squeezing-Averse Virtual Try-On via Sequential Deformation}}} \\
    \vspace*{0.1cm}
    \bf{\Large{Sang-Heon Shim, Jiwoo Chung, Jae-Pil Heo\footnotemark[1]}} \\
    \vspace*{0.1cm}
    \large{\normalfont{Sungkyunkwan University}} \\
    \large{\normalfont{\{ekzmwww, wldn0202, jaepilheo\}@skku.edu}}
    \vspace*{1.6cm}
\end{center}]

\footnotetext{$^\ast$Corresponding author}

\input{7_supp}

\end{document}

%% file: 7_supp.tex
In this supplementary material, we first provide implementation details of our try-on condition generator, and then present the additional qualitative results.
Lastly, we compare the number of learnable parameters and inference times between methods.
Since there are many qualitative results, we recommend not to print this material.

\section{Implementation Details}
We implement our proposed method on top of the official PyTorch implementation of HR-VITON~\cite{lee2022hrviton}.
Our implementation details are the same with HR-VITON.
Specifically, we set the batch sizes to $8$ when training our try-on condition generator.
The learning rate is set to $0.0002$. 
We train our model until $100,000$ iterations.
The training and inference strategies are also the same with HR-VITON.
During the training phase, our try-on condition generator predicts the segmentation map~($\hat{S}$), the warped clothes~($\hat{I}_c$), and the warped clothes mask~($\hat{S}_c$), each of which has $256 \times 192$ resolution.
In the inference stage, the appearance flow map~($F$) and $\hat{S}$ are upscaled to $1024 \times 768$ resolution.
Thus, $\hat{S}$, $\hat{I}_c$, and $\hat{S}_c$ with $1024 \times 768$ resolution are given to our try-on image generator.

\begin{figure}[t]
	\centering
    \includegraphics[width=\linewidth]{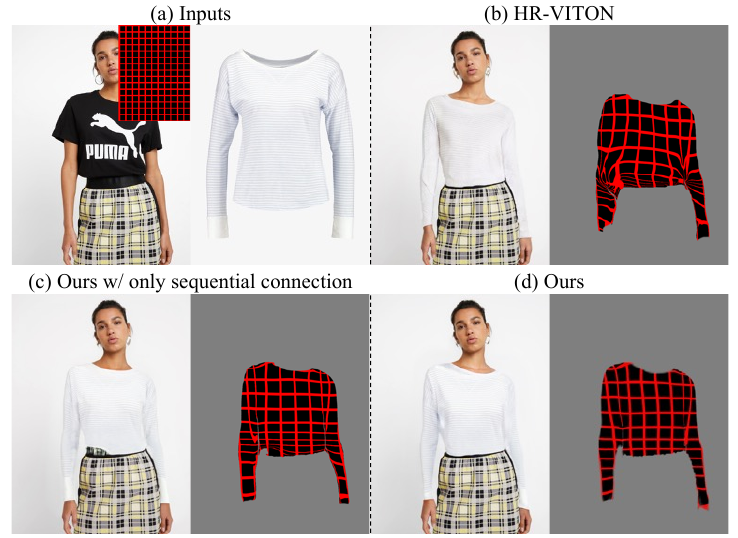}    
    \caption{Visualization of a try-on sample and a warped grid. (b) The warped grid of HR-VITON is squeezed nearby the sleeve and the waist. (c) Our ablated model produces a warped grid without squeezing around the sleeve, but the grid is progressively narrowed around the waist. (d) Ours full model deforms a grid without any squeezing problem.}
	\label{fig_warped_grid}
\end{figure}

\section{More Qualitative Results}

\paragraph{Visualization of warped grid.} 
Given a grid image, Fig.~\ref{fig_warped_grid} shows a comparison of warped grid between different methods.
The warped grid of HR-VITON is squeezed nearby the arm and waist.
On the other hands, ours w/ sequential connection does not have a squeezed grid on the arm.
However, we observe that the area of grid gradually shrinks nearby the waist.
This result implies that the model decided to deform clothes with preserving the whole contents with squeezed textures.
Lastly, our full model successfully removes the partial contents of clothes and has a consistent interval between grids.
These results confirm that our proposed methods successfully address our raised two types of squeezing artifacts.

\begin{figure*}[t!]
    \centering
     \includegraphics[width=\textwidth]{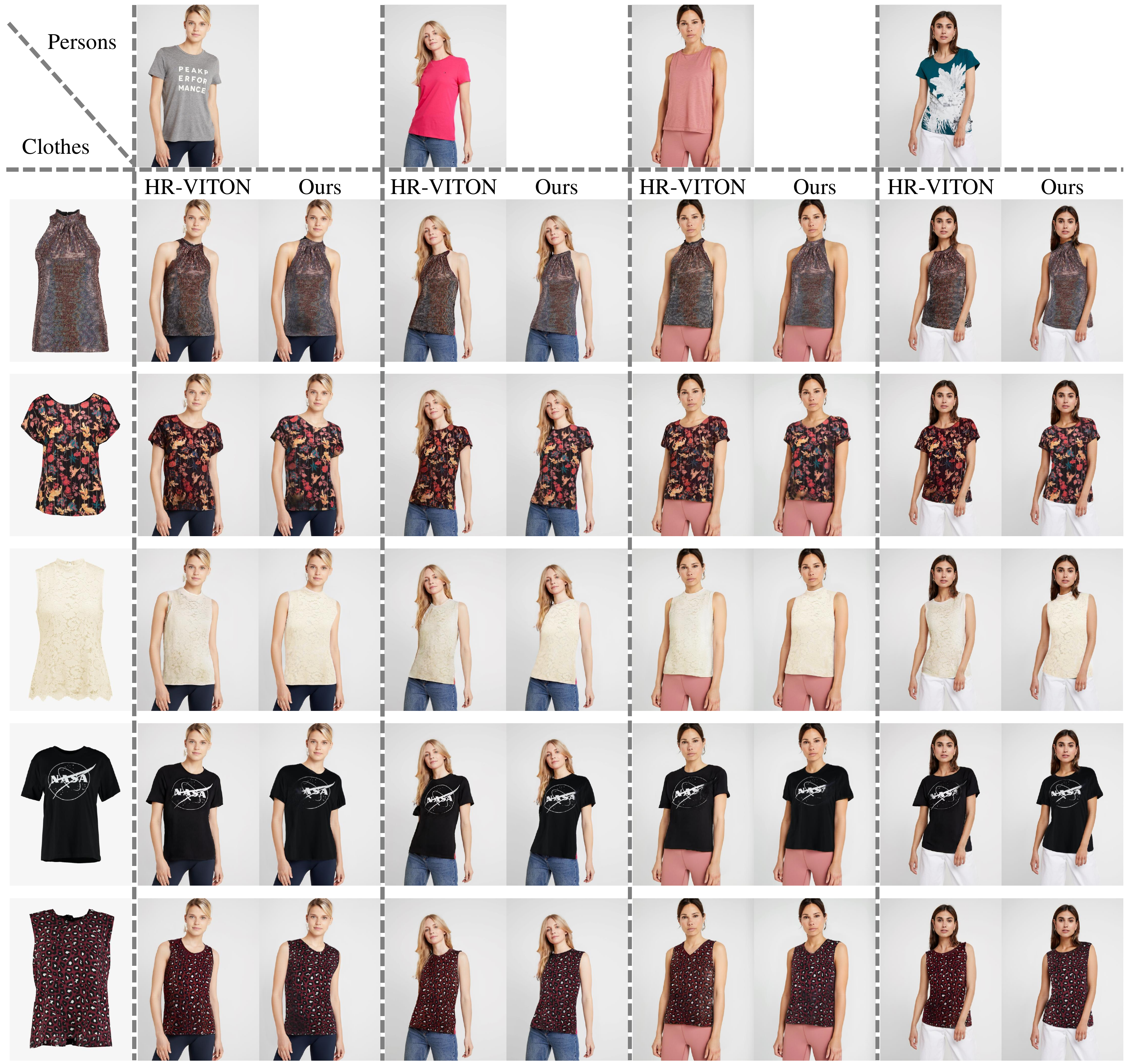}
     \caption{Qualitative comparison between HR-VITON~\cite{lee2022hrviton} and ours. The input pair is the person with a tucked-out shirts style and the clothes with short- or no sleeves which are not our main concern. For those pairs, the visual quality of generated samples is indistinguishable between the baseline and ours. These results show that our proposed method does not degrade the performance even for the configuration different from our main subject in this paper.
     \label{fig_supp_causal}
     }
\end{figure*}

\paragraph{Generated samples for persons with tucked-out shirts style and clothes with short- or no sleeves} are reported in Fig.~\ref{fig_supp_causal}.
As seen in the results, the visual quality of generated samples is indistinguishable between ours and HR-VITON~\cite{lee2022hrviton}.
This shows that our proposed method does not degrade the visual quality
even for the configuration different from our main subject, 
clothes with long sleeves or a person with a tucked-in shirts style.

\begin{figure*}[t!]
    \centering
     \includegraphics[width=\textwidth]{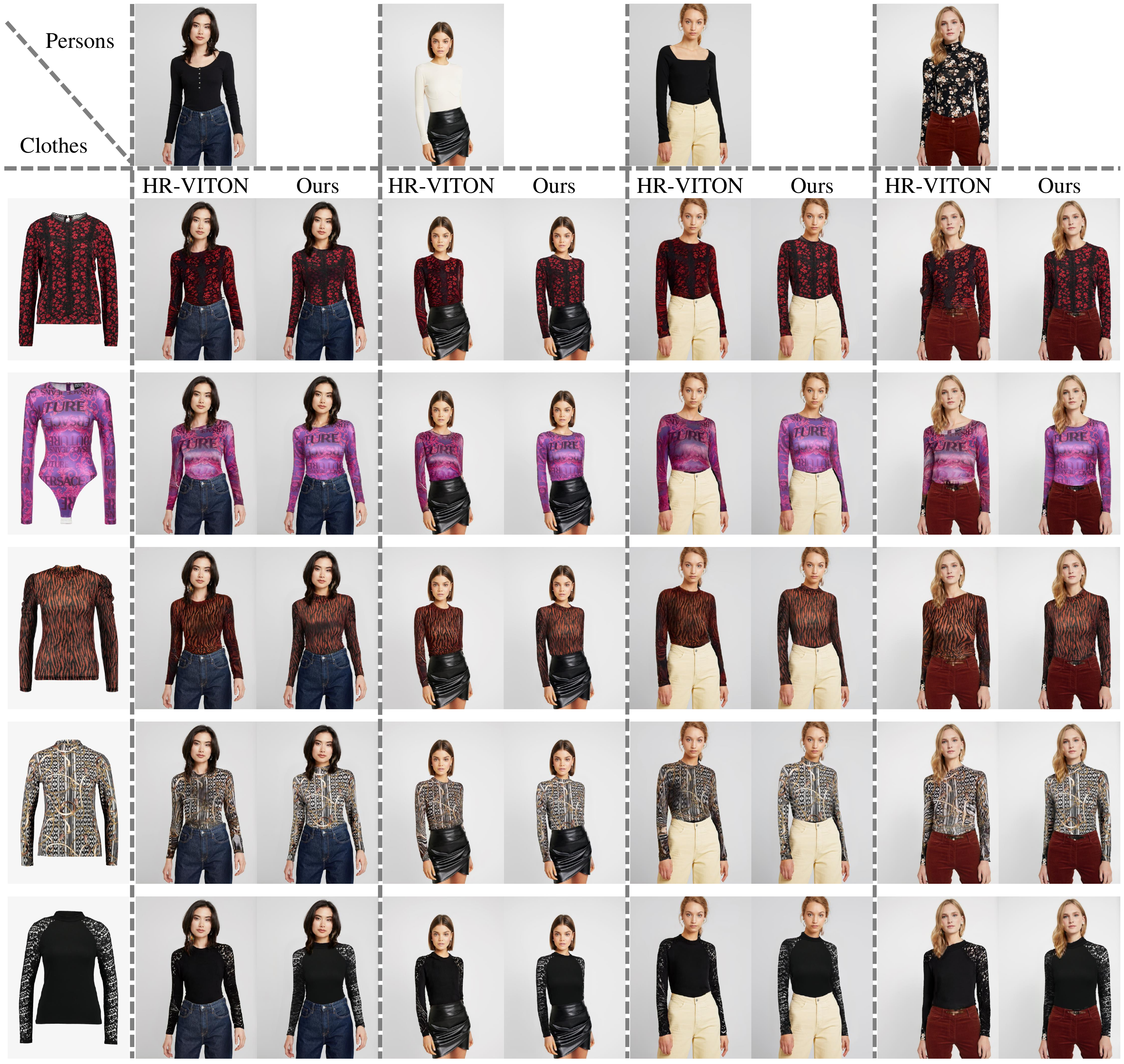}
     \caption{Qualitative comparison between HR-VITON~\cite{lee2022hrviton} and ours. We recommend to look into the arm. Most generated samples from HR-VITON have squeezed textures on arm, whereas the synthesized samples from our model have the unique clothing texture on arm.
     }
     \label{fig_supp_longsleeve}
\end{figure*}

\begin{figure*}[t!]
    \centering
     \includegraphics[width=0.85\textwidth]{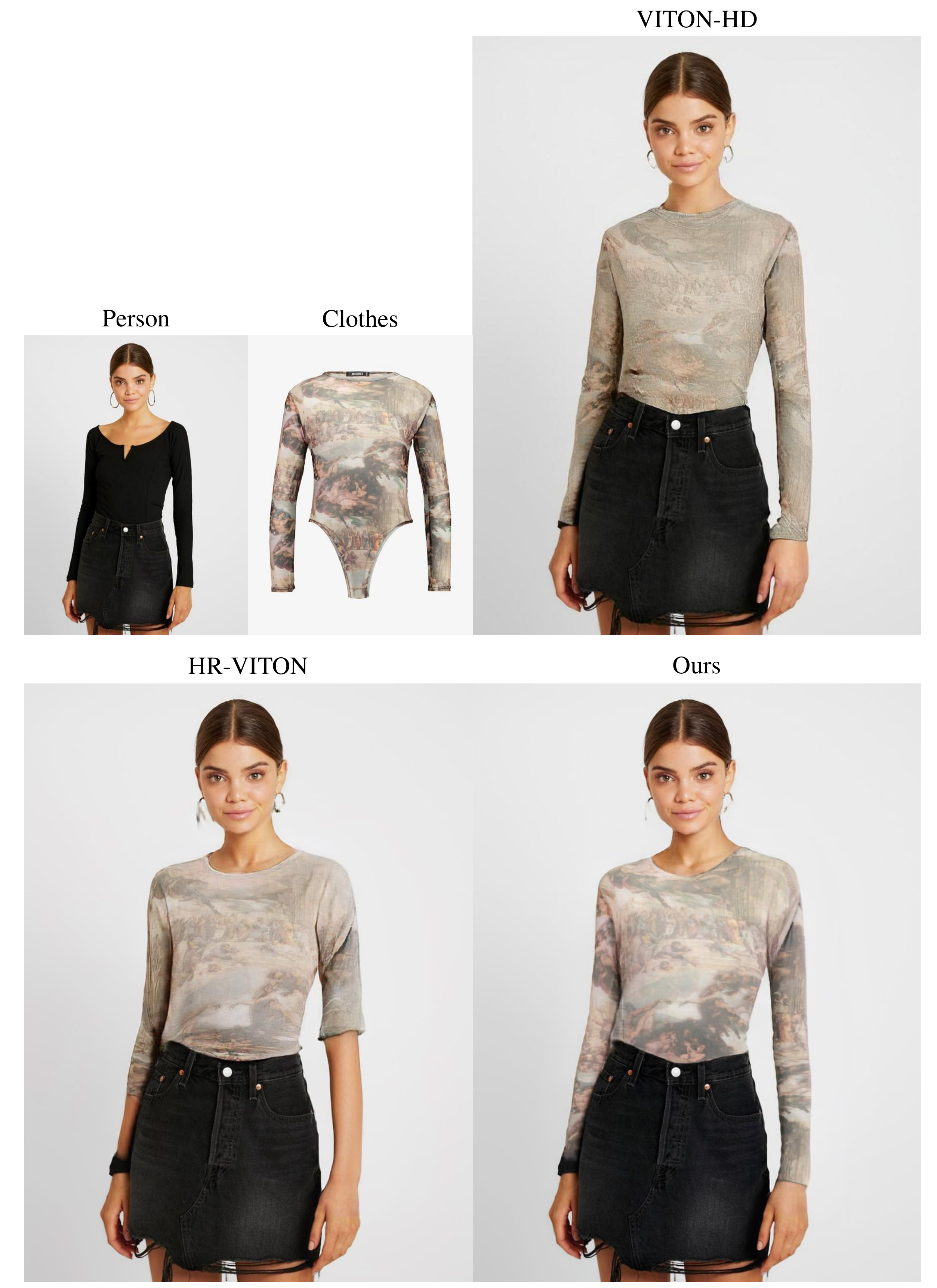}
     \caption{Qualitative comparison between VITON-HD~\cite{choi2021viton}, HR-VITON~\cite{lee2022hrviton}, and ours. We recommend to look into the arm.
     }
     \label{fig_supp_longsleeve_HR_0}
\end{figure*}

\begin{figure*}[t!]
    \centering
     \includegraphics[width=0.85\textwidth]{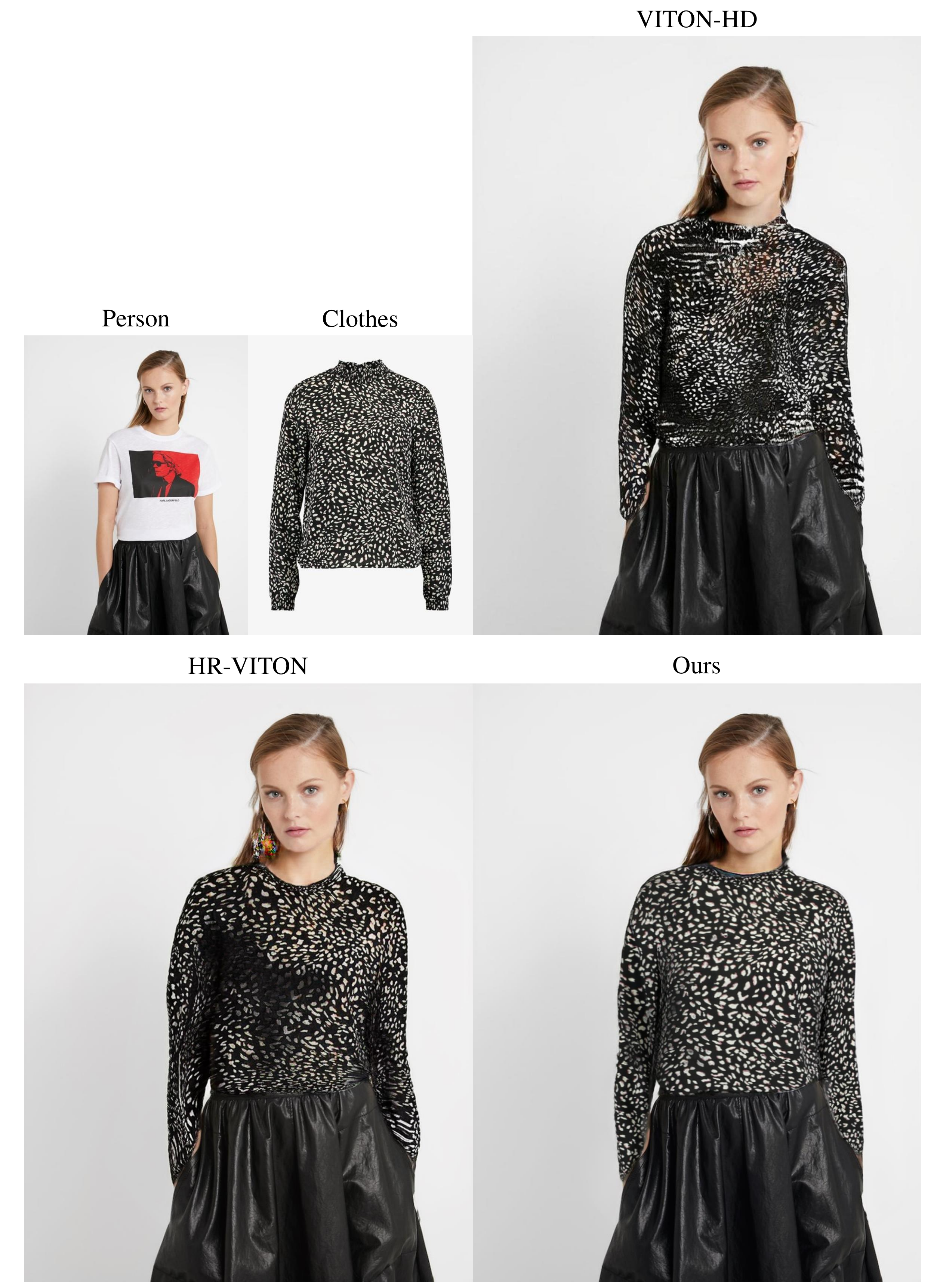}
     \caption{Qualitative comparison between VITON-HD~\cite{choi2021viton}, HR-VITON~\cite{lee2022hrviton}, and ours. We recommend to look into the arm.
     }
     \label{fig_supp_longsleeve_HR_1}
\end{figure*}

\begin{figure*}[t!]
    \centering
     \includegraphics[width=0.85\textwidth]{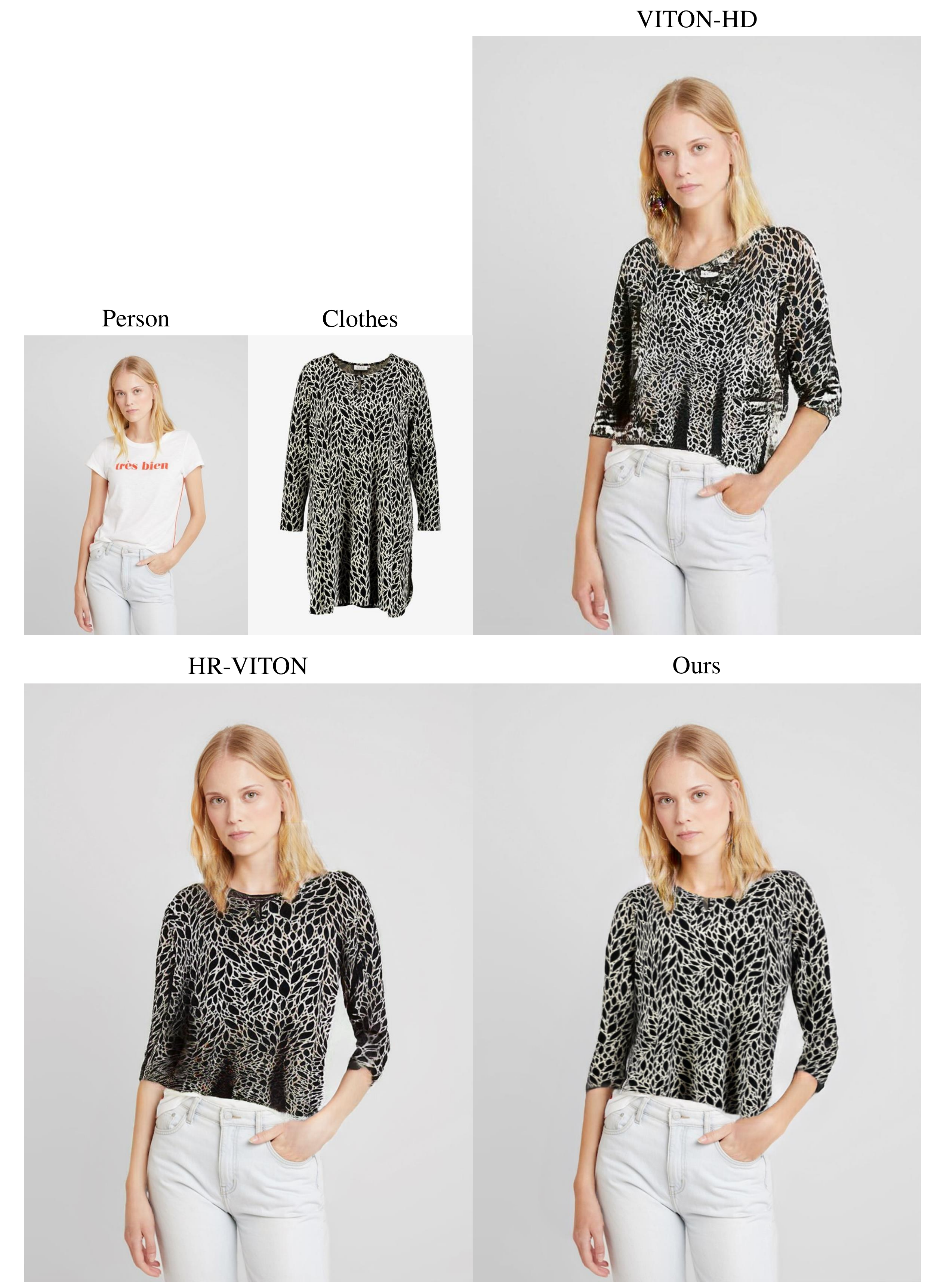}
     \caption{Qualitative comparison between VITON-HD~\cite{choi2021viton}, HR-VITON~\cite{lee2022hrviton}, and ours. We recommend to look into the arm.
     }
     \label{fig_supp_longsleeve_HR_2}
\end{figure*}

\begin{figure*}[t!]
    \centering
     \includegraphics[width=0.85\textwidth]{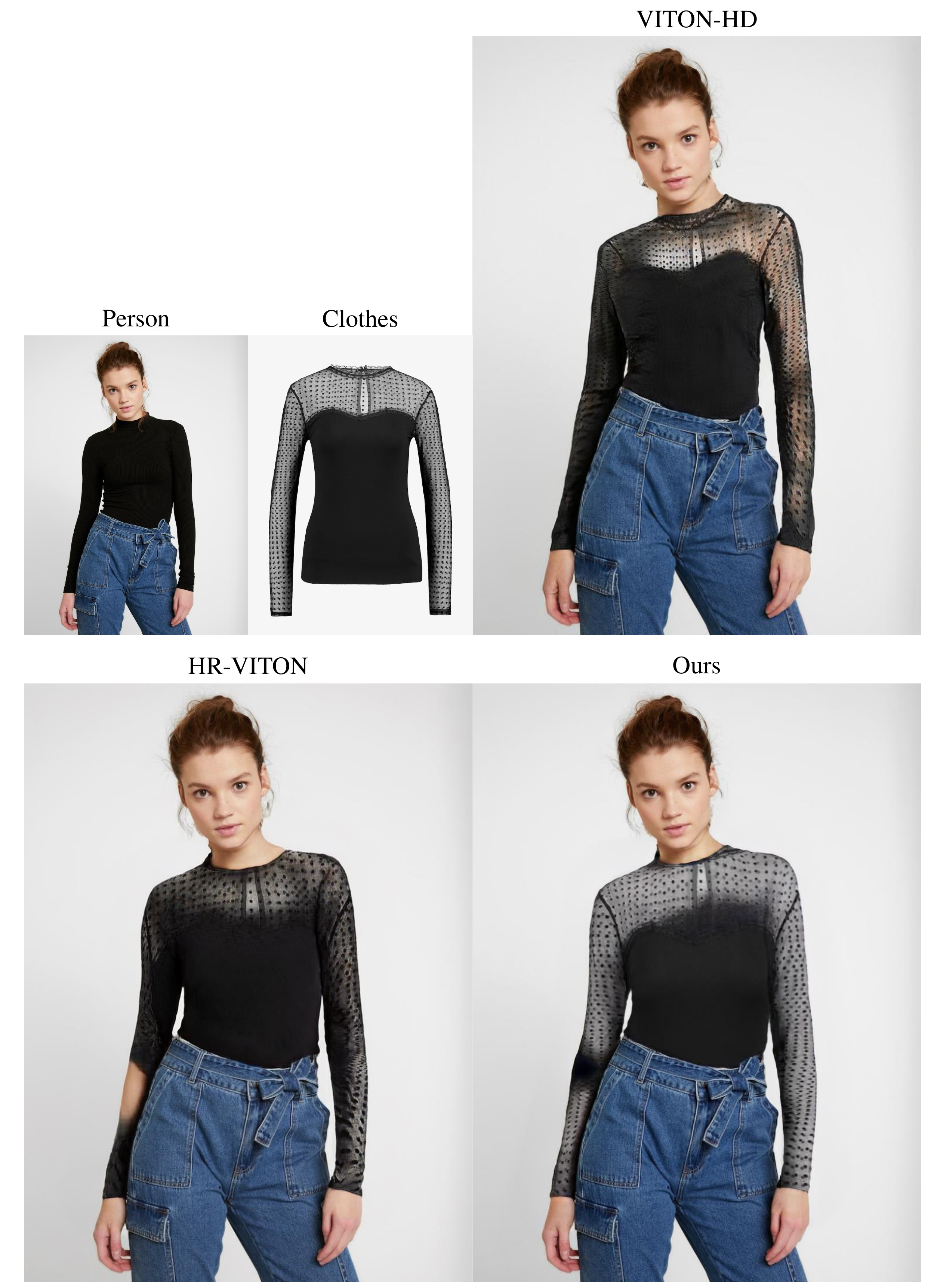}
     \caption{Qualitative comparison between VITON-HD~\cite{choi2021viton}, HR-VITON~\cite{lee2022hrviton}, and ours. We recommend to look into the arm.
     }
     \label{fig_supp_longsleeve_HR_3}
\end{figure*}

\begin{figure*}[t!]
    \centering
     \includegraphics[width=0.85\textwidth]{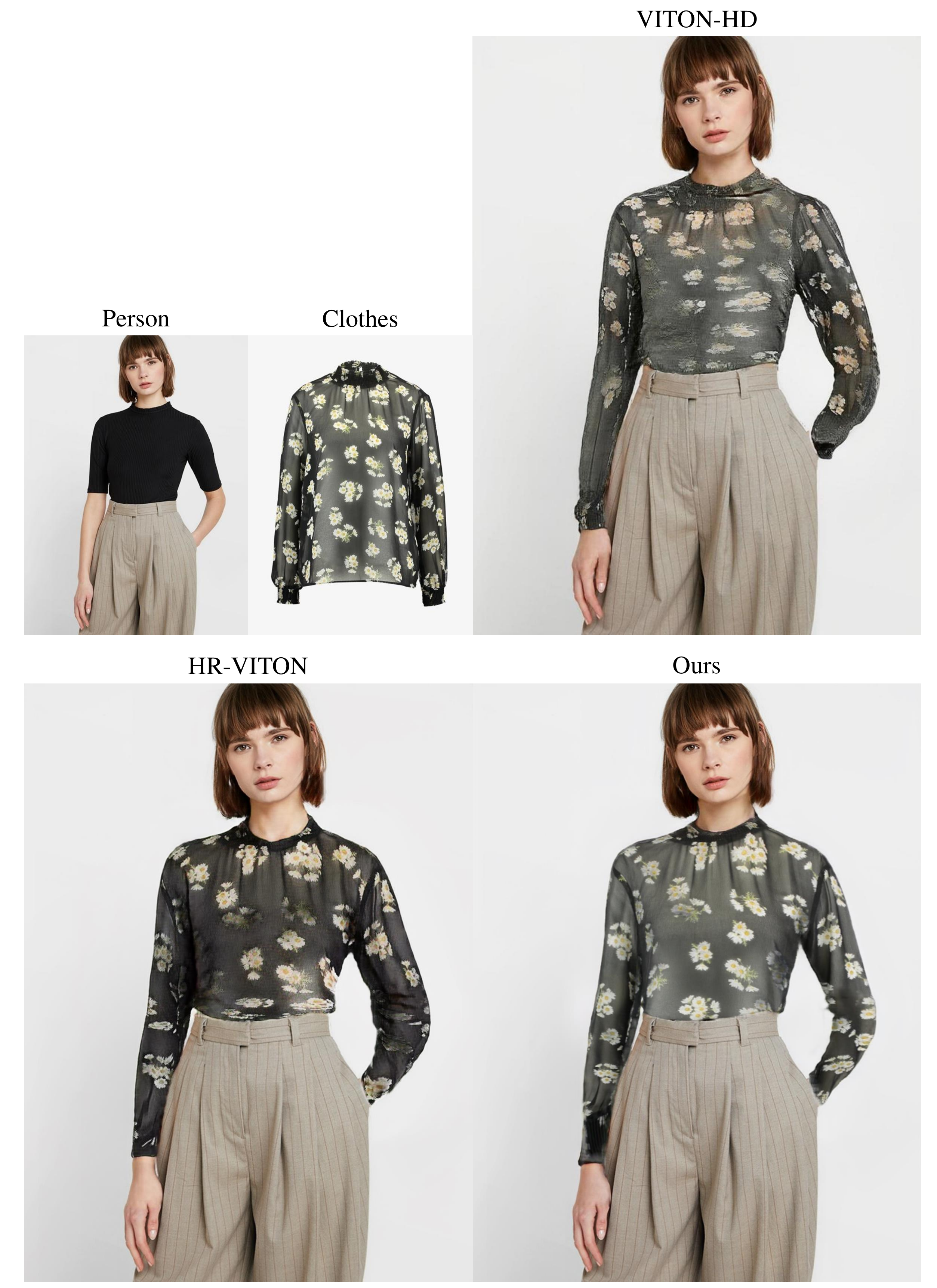}
     \caption{Qualitative comparison between VITON-HD~\cite{choi2021viton}, HR-VITON~\cite{lee2022hrviton}, and ours. We recommend to look into the arm.
     }
     \label{fig_supp_longsleeve_HR_4}
\end{figure*}

\begin{figure*}[t!]
    \centering
     \includegraphics[width=0.85\textwidth]{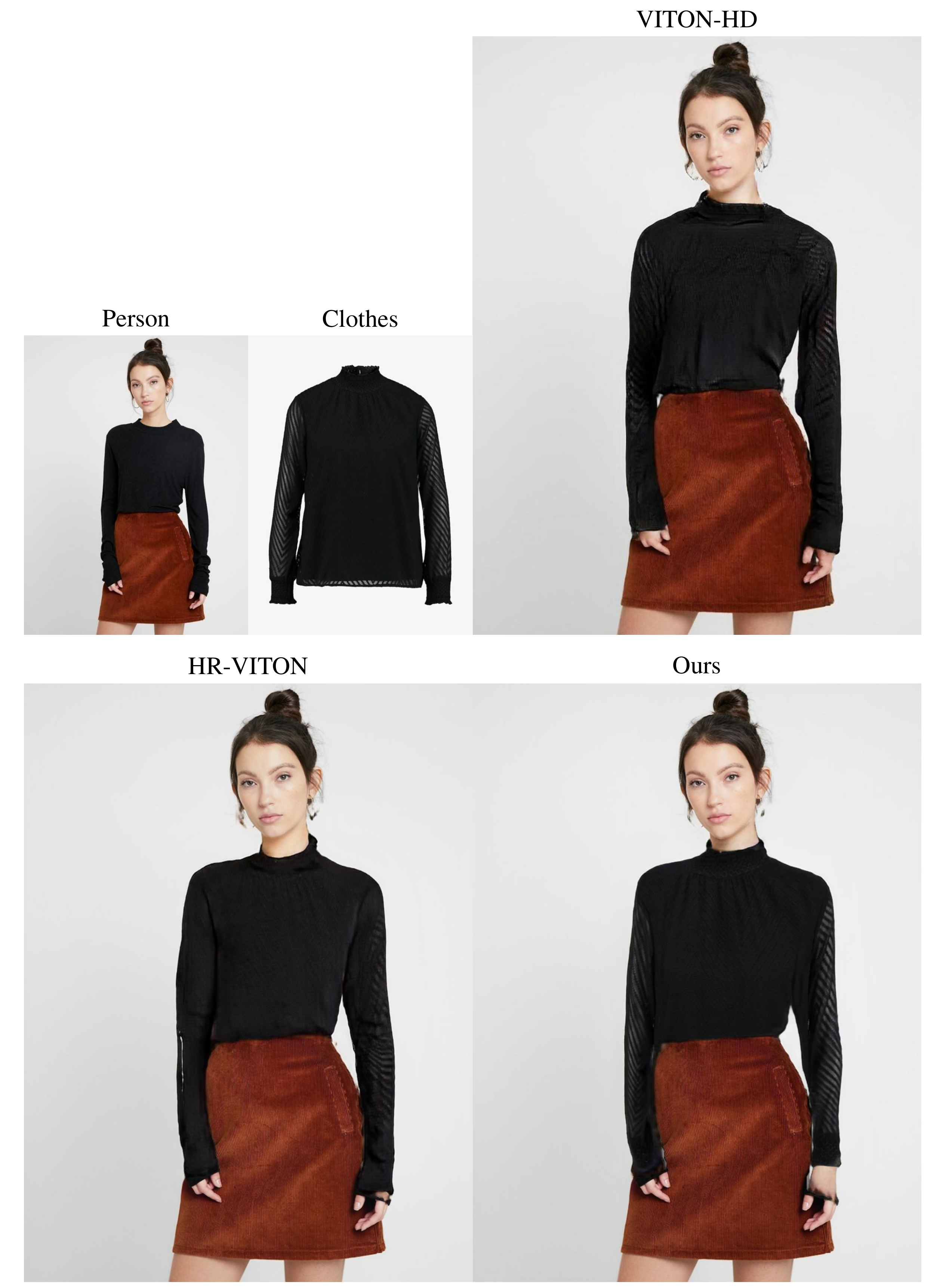}
     \caption{Qualitative comparison between VITON-HD~\cite{choi2021viton}, HR-VITON~\cite{lee2022hrviton}, and ours. We recommend to look into the arm.
     }
     \label{fig_supp_longsleeve_HR_5}
\end{figure*}

\paragraph{Sleeve-squeezing artifact.}
Fig.~\ref{fig_supp_longsleeve} shows the comparison of generated samples when clothes with long sleeves are given.
In most tested cases, HR-VITON suffers from the sleeve-squeezing artifact.
On the other hand, our method successfully generates the samples by preserving the unique textures of clothes on arms.
Additional high-resolution qualitative results are available in Fig.~\ref{fig_supp_longsleeve_HR_0} -- \ref{fig_supp_longsleeve_HR_5}.

\begin{figure*}[t!]
    \centering
     \includegraphics[width=\textwidth]{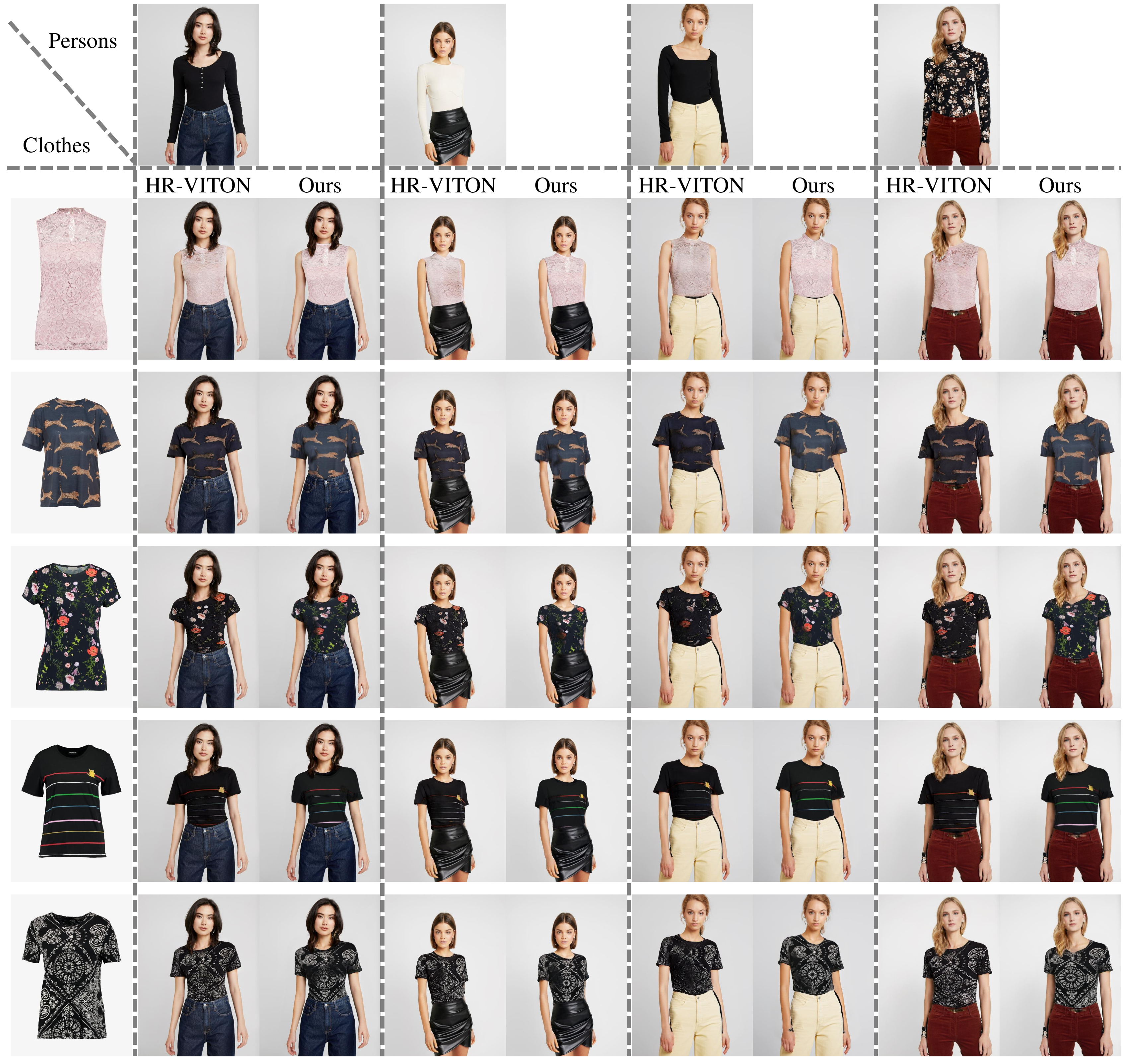}
     \caption{Qualitative comparison between HR-VITON~\cite{lee2022hrviton} and ours. We recommend to look into the waist. The baseline produces the samples with preserving the whole contents with squeezed textures, while our model generates the samples without squeezed textures nearby the waist since our model is able to erase the partial contents of clothes.
     }
     \label{fig_supp_tucked_in}
\end{figure*}

\begin{figure*}[t!]
    \centering
     \includegraphics[width=0.85\textwidth]{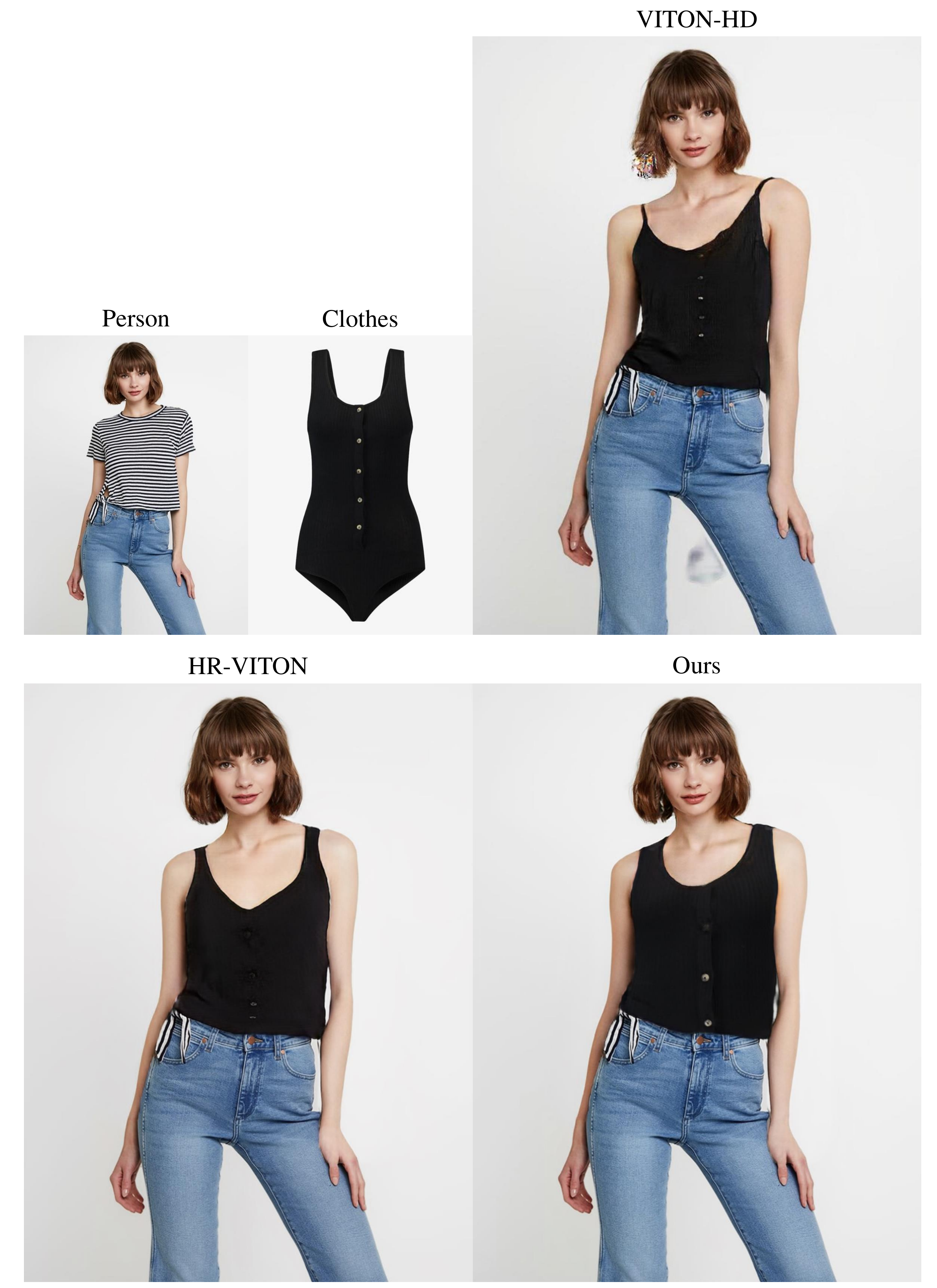}
     \caption{Qualitative comparison between VITON-HD~\cite{choi2021viton}, HR-VITON~\cite{lee2022hrviton}, and ours. We recommend to look into the waist.
     }
     \label{fig_supp_tucked_in_HR_6}
\end{figure*}

\begin{figure*}[t!]
    \centering
     \includegraphics[width=0.85\textwidth]{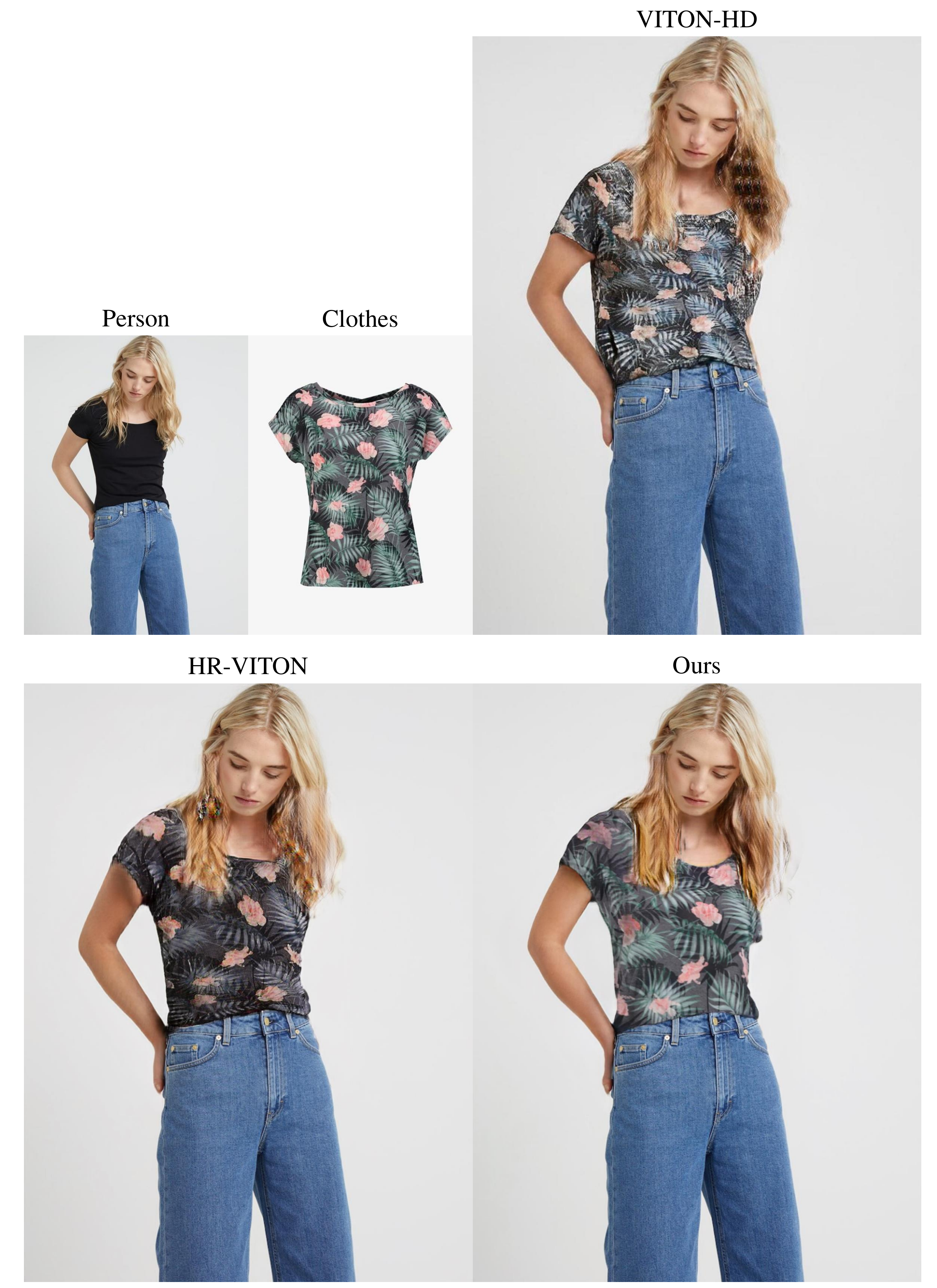}
     \caption{Qualitative comparison between VITON-HD~\cite{choi2021viton}, HR-VITON~\cite{lee2022hrviton}, and ours. We recommend to look into the waist.
     }
     \label{fig_supp_tucked_in_HR_7}
\end{figure*}

\begin{figure*}[t!]
    \centering
     \includegraphics[width=0.85\textwidth]{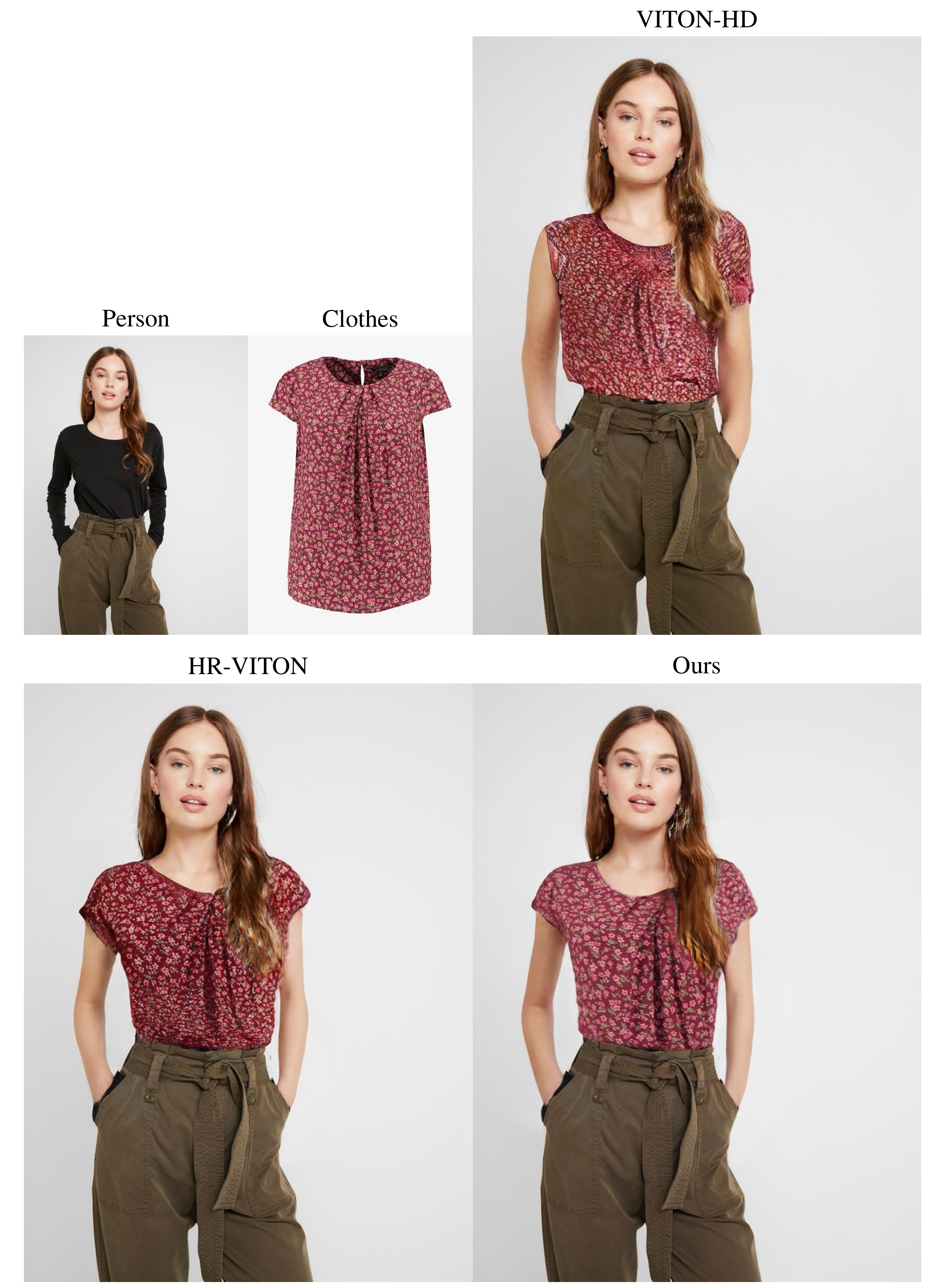}
     \caption{Qualitative comparison between VITON-HD~\cite{choi2021viton}, HR-VITON~\cite{lee2022hrviton}, and ours. We recommend to look into the waist.
     }
     \label{fig_supp_tucked_in_HR_8}
\end{figure*}

\begin{figure*}[t!]
    \centering
     \includegraphics[width=0.85\textwidth]{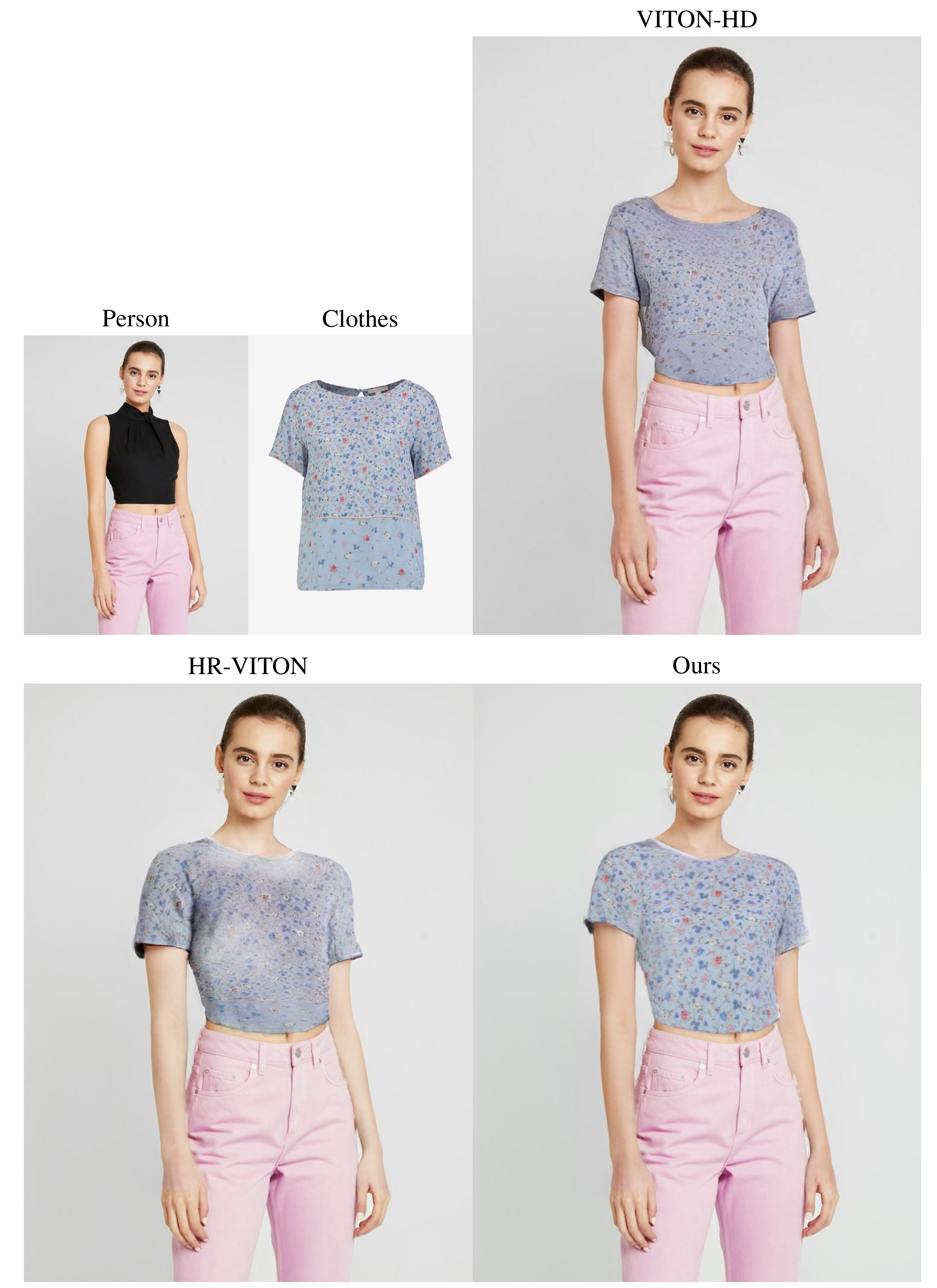}
     \caption{Qualitative comparison between VITON-HD~\cite{choi2021viton}, HR-VITON~\cite{lee2022hrviton}, and ours. We recommend to look into the waist.
     }
     \label{fig_supp_tucked_in_HR_9}
\end{figure*}

\begin{figure*}[t!]
    \centering
     \includegraphics[width=0.85\textwidth]{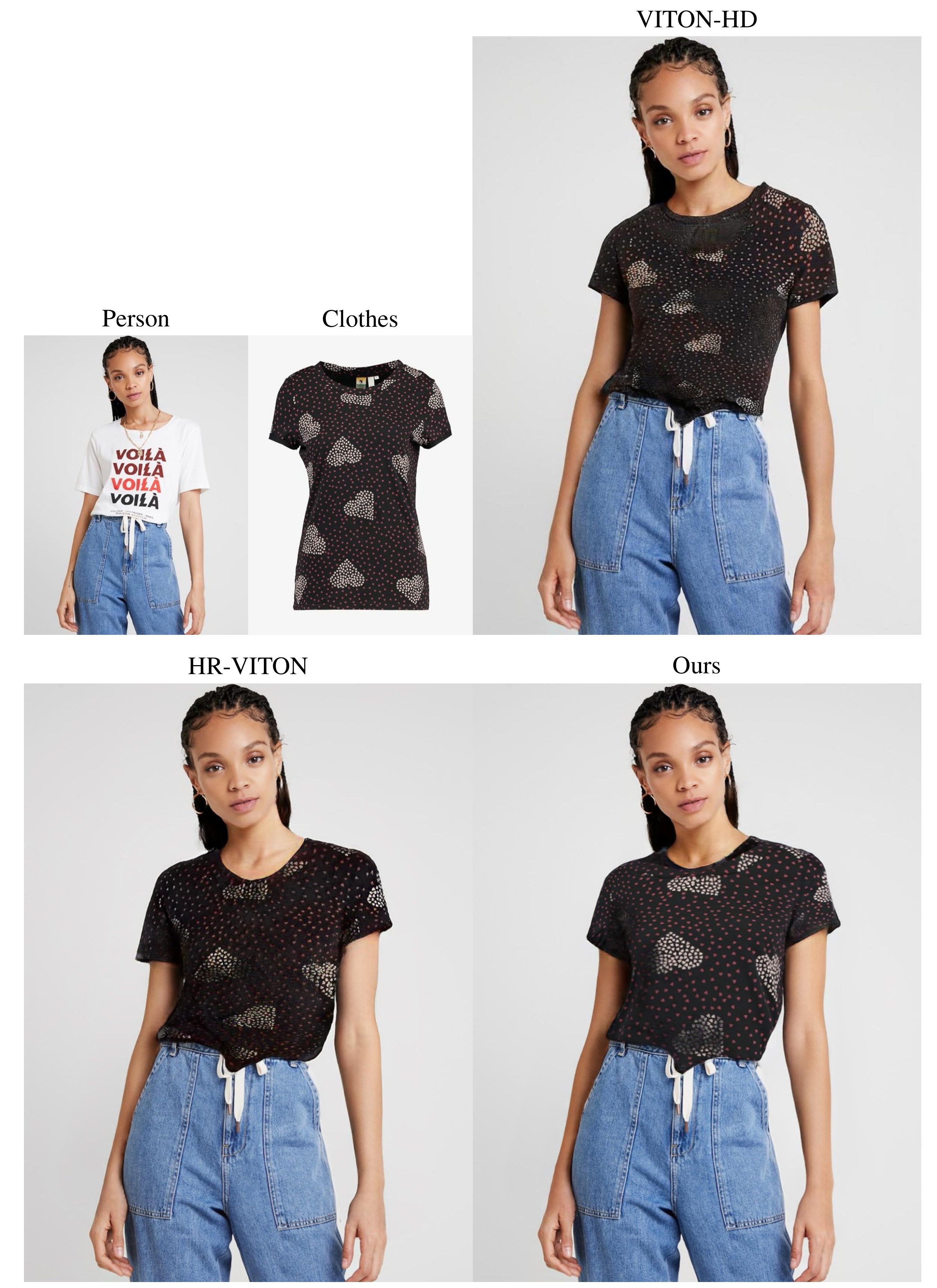}
     \caption{Qualitative comparison between VITON-HD~\cite{choi2021viton}, HR-VITON~\cite{lee2022hrviton}, and ours. We recommend to look into the waist.
     }
     \label{fig_supp_tucked_in_HR_10}
\end{figure*}

\begin{figure*}[t!]
    \centering
     \includegraphics[width=0.85\textwidth]{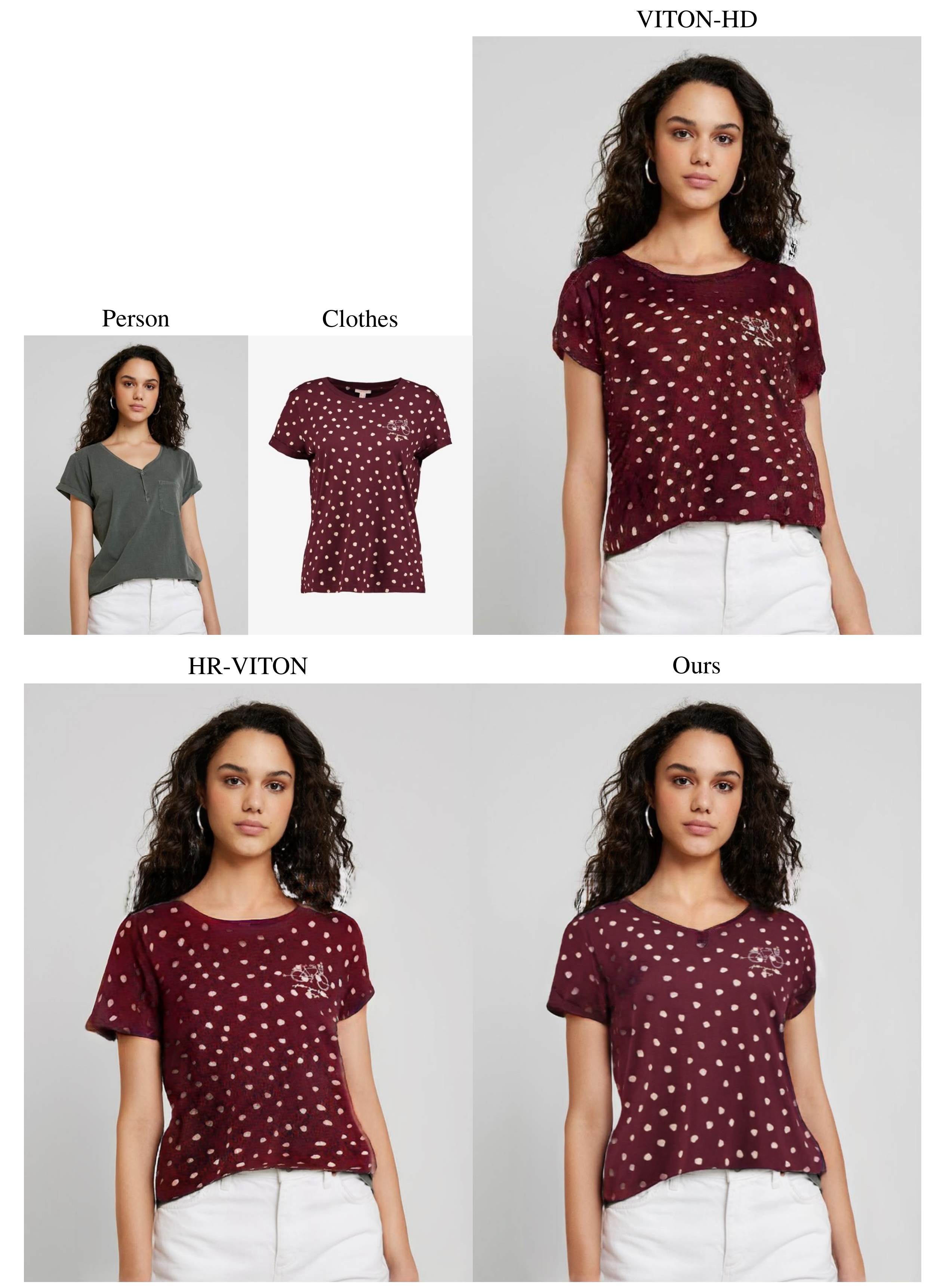}
     \caption{Qualitative comparison between VITON-HD~\cite{choi2021viton}, HR-VITON~\cite{lee2022hrviton}, and ours. We recommend to look into the waist.
     }
     \label{fig_supp_tucked_in_HR_11}
\end{figure*}

\paragraph{Waist-squeezing artifact.}
We also report the qualitative comparison with respect to waist-squeezing artifacts.
As shown in Fig.~\ref{fig_supp_tucked_in}, HR-VITON generates the samples with preserving the whole contents with a squeezed texture nearby a waist.
On the other hand, our method produces the samples without squeezed texture around the waist, since our model is able to erase the partial contents of clothes.
More high-resolution qualitative results are presented in Fig.~\ref{fig_supp_tucked_in_HR_6} -- \ref{fig_supp_tucked_in_HR_11}.

\begin{table}[t]
	\centering
	\begin{tabular}{l|c|c}
		\toprule
		Methods  & \#Params of Gs & Frames per second \\
		\midrule
		\midrule
        VITON-HD  & 154.07M & N/A \\
        HR-VITON  & \textbf{147.91M} &  \textbf{2.49} \\
        Ours & 148.14M & \textbf{2.49} \\
		\bottomrule
	\end{tabular}
    \caption{Number of parameters and inference time.
    }
	\label{tb_computational_costs}
\end{table}

\section{Number of Parameters and Inference Time}
\paragraph{Number of learnable parameters} are reported in Table.~\ref{tb_computational_costs}. Since our methods only require an additional 3D convolution operation at the last fusion block, the increment of learnable parameters is $0.23$ million than HR-VITON.
It clarifies that our improvement is not caused by increasing the model capacity. 

\paragraph{Inference times} are compared by measuring the frames-per-second~(FPS) of HR-VITON and Ours. Note that, the batch size is set to $1$ when measuring times. As reported in Table.~\ref{tb_computational_costs}, the inference times are nearly identical (2.49 FPS) on a single A6000 GPU.

